\PassOptionsToPackage{table,dvipsnames}{xcolor}

\documentclass[11pt,a4paper]{article}
\usepackage{times,latexsym}
\usepackage{url}
\usepackage[T1]{fontenc}

\usepackage[acceptedWithA]{tacl2021v1}

\usepackage{xspace,mfirstuc,tabulary}

\newif\iftaclinstructions
\taclinstructionsfalse 
\iftaclinstructions
\renewcommand{\confidential}{}
\renewcommand{\anonsubtext}{(No author info supplied here, for consistency with
TACL-submission anonymization requirements)}
\newcommand{\instr}
\fi

\iftaclpubformat 

\else

\fi

\usepackage{microtype}
\usepackage{hyperref}
\usepackage{url}
\usepackage{booktabs}
\usepackage{graphicx}
\usepackage{todonotes}
\usepackage{arydshln}
\usepackage{pifont}
\usepackage{array}
\renewcommand{\arraystretch}{0.8}
\usepackage{inconsolata}
\usepackage{adjustbox}
\usepackage{soul}
\usepackage{amsmath}
\usepackage{changepage}
\usepackage[inline]{enumitem}
\usepackage{makecell}
\usepackage{colortbl}
\usepackage{mdframed}
\usepackage{soul}
\usepackage{tablefootnote}

\newcommand{\hlpink}[1]{{\sethlcolor{pink}\hl{#1}}}

\newcommand\dolomites{\raisebox{-1pt}{\includegraphics[width=1.75em]{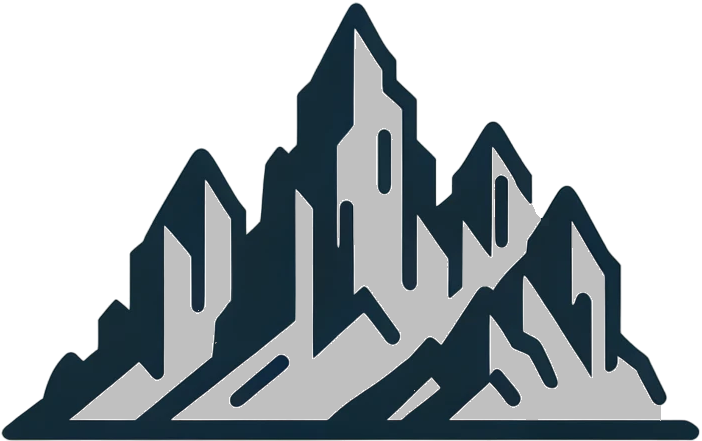}}}
\newcommand\pennemoji{\raisebox{-2pt}{\includegraphics[width=0.7em]{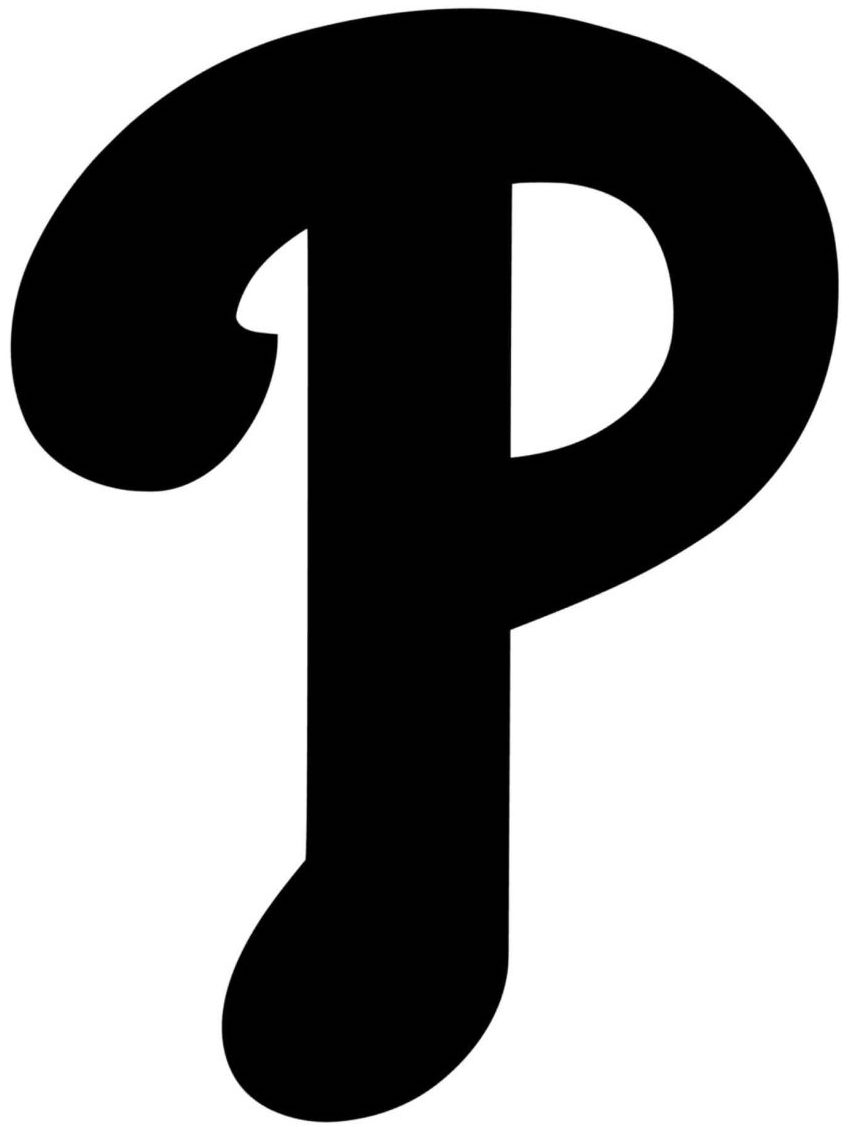}}}
\newcommand\googleemoji{\raisebox{-2pt}{\includegraphics[width=0.9em]{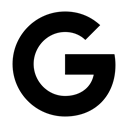}}}

\title{\dolomites{} \textsc{Dolomites}: Domain-Specific Long-Form Methodical Tasks}

\author{Chaitanya Malaviya\textsuperscript{\pennemoji}\thanks{\,\,\,Work done at Google DeepMind.},
    Priyanka Agrawal\textsuperscript{\googleemoji},
    Kuzman Ganchev\textsuperscript{\googleemoji},\\
    \textbf{Pranesh Srinivasan}\textsuperscript{\googleemoji},
    \textbf{Fantine Huot}\textsuperscript{\googleemoji},
    \textbf{Jonathan Berant}\textsuperscript{\googleemoji},
    \textbf{Mark Yatskar}\textsuperscript{\pennemoji},\\
    \textbf{Dipanjan Das}\textsuperscript{\googleemoji},
    \textbf{Mirella Lapata}\textsuperscript{\googleemoji},
    \textbf{Chris Alberti}\textsuperscript{\googleemoji}
    \vspace{0.1in} \\
    \textsuperscript{\pennemoji}University of Pennsylvania
    \textsuperscript{\googleemoji}Google DeepMind\\
    {\tt cmalaviy@seas.upenn.edu}\\
}

\date{}

\begin{document}
\maketitle
\begin{abstract}
Experts in various fields routinely perform methodical writing tasks
  to plan, organize, and report their work. From a clinician writing a
  differential diagnosis for a patient, to a teacher writing a lesson
  plan for students, these tasks are pervasive, requiring to
  \emph{methodically} generate structured long-form output for a given
  input. We develop a typology of methodical tasks structured in the
  form of a task objective, procedure, input, and output, and
  introduce DoLoMiTes, a novel benchmark with specifications for
  519~such tasks elicited from hundreds of experts from across
  25~fields. Our benchmark further contains specific instantiations of
  methodical tasks with concrete input and output examples (1,857 in
  total) which we obtain by collecting expert revisions of up to 10
  model-generated examples of each task. We use these examples to
  evaluate contemporary language models, highlighting that automating
  methodical tasks is a challenging long-form generation problem, as
  it requires performing complex inferences, while drawing upon the
  given context as well as domain knowledge.
  Our dataset is available at \url{https://dolomites-benchmark.github.io/}.
\end{abstract}

\section{Introduction}


\begin{figure}[t!]
    \centering
    \includegraphics[width=\columnwidth]{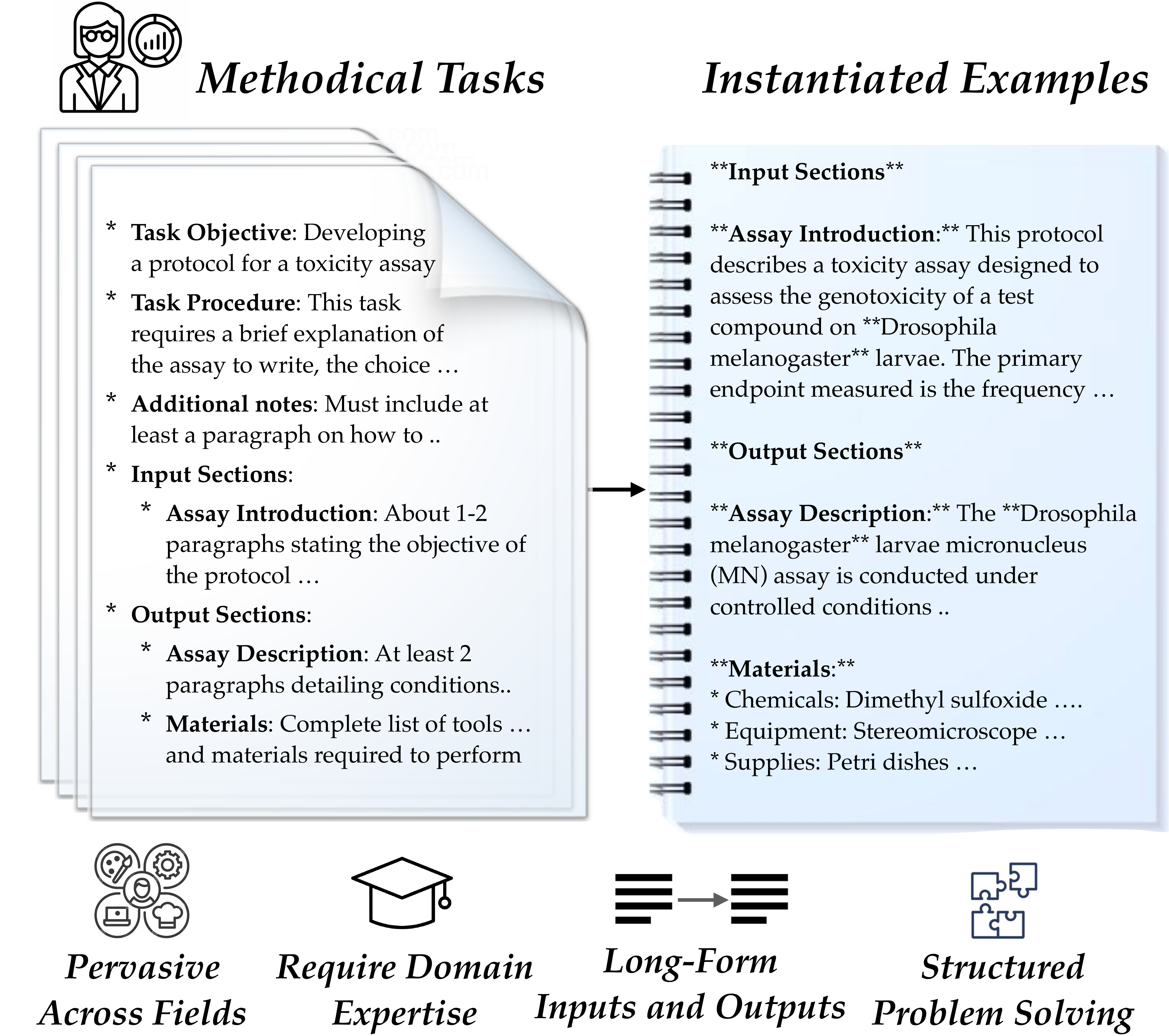}
    \caption{\textsc{Dolomites} contains descriptions of 519 methodical tasks elicited from domain experts across various fields. We instantiate these tasks with examples that contain plausible inputs and outputs, formulating a challenging long-form generation problem that requires domain expertise and structured problem-solving.}
    \label{fig:overview}
\end{figure}

Experts in various fields regularly use writing as a means for
planning, organizing, and sharing their work. For instance, a teacher
might draft a lesson plan for what they would like to teach in their
next class, and a lawyer might draft a patent application for an
invention. Experts generally follow a consistent and methodical
approach to conduct these writing tasks. In the lesson plan example, a
teacher would know the lesson objectives, format and profile of the
class, and would produce a plan with the topics to be covered and
activities to improve learning. Importantly, the teacher follows a
systematic procedure to write this lesson plan, using their expertise
and what they know about the current context (e.g., the class
profile).


\begin{figure*}[ht!]
    \centering
    \includegraphics[width=2\columnwidth]{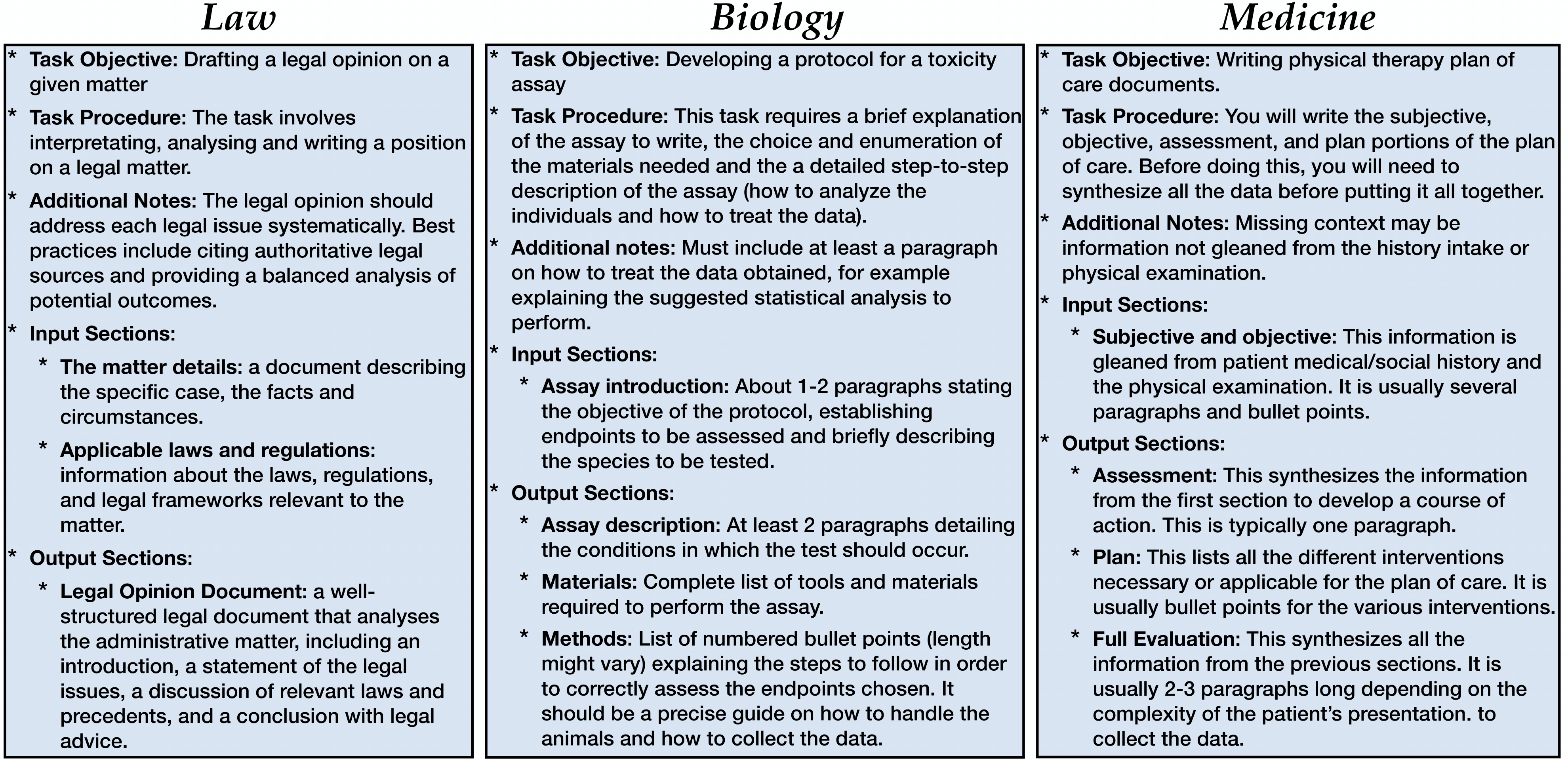}
    \caption{A sample of methodical tasks from law, biology and medicine in \textsc{Dolomites}. Each task in \textsc{Dolomites} follows a standard template, containing a task objective, task procedure, additional notes about the task, and finally, input sections that are usually expected for the task, and output sections that need to be produced as part of the task. These tasks are instantiated with \textit{examples} that represent plausible inputs and outputs for the task (section \ref{sec:example_collection}).}
    \vspace{-10pt}
    \label{fig:sample_tasks}
\end{figure*}

Across fields, from law to visual arts and engineering, experts
accomplish on a regular basis such \textit{methodical} tasks, i.e.,
writing tasks which loosely follow a standard template for what is
usually given as input and what is required from the output. These
tasks often follow a structured and consistent procedure as they are
performed regularly and tend to be fairly time-consuming, taking from
a few hours to several days (see Figure~\ref{fig:overview}).  As large
language models become more capable and widely accessible to a more
sophisticated set of users \citep{owens2023nature,mollick2023using,lee2023benefits,birhane2023,mollick2023using,frankenreiter2022natural,demszky2023using,wang2023scientific},
they hold great potential for assisting experts with methodical
writing tasks, increasing their efficiency and allowing them to
focus on complex problem-solving activities
\citep{noy2023experimental}.



Given their potential for assisting experts, it would be beneficial to
evaluate language models on a realistic set of methodical writing
tasks. However, we currently do not have benchmarks that contain a
typology of such tasks.
The most natural source for such data would be query logs
\citep{nguyen2016ms,kwiatkowski-etal-2019-natural} or chat histories
\citep{zhao2023inthe}. However, these data sources do not specifically
reflect domain-specific use cases and do not allow us to study
specific use cases in a controlled manner.





In this work, we bridge this gap by eliciting 519 methodical task
descriptions (see a few examples in Figure~\ref{fig:sample_tasks})
from 266 experts across 25 different fields
(Section~\ref{sec:task_collection}). These writing tasks are formatted
in a standard way, with a task procedure, input, and output. Further
analysis with an independent group of experts reveals that they are
indeed plausible ($\sim$76\% of them are likely to be conducted by an
expert on a regular basis) and most experts ($\sim$63\%) would find it
useful if they could use a capable AI model as a writing assistant
(Section~\ref{sec:task_validation}). Our tasks serve as the first
collection of realistic use cases of experts spanning multiple
domains.

To evaluate the ability of existing models to assist experts with
these tasks, we collect \textit{examples} (see
Figure~\ref{fig:overview}), where we instantiate each task with
plausible inputs and outputs
(Section~\ref{sec:example_collection}). Examples are created
semi-automatically: we first retrieve web documents that could
potentially serve as samples of the task, and then generate an example
using a language model based on the retrieved web documents. These
examples are then significantly post-edited by the same experts (who
contributed the task) for improving adherence to the task description, factual correctness,
and level of detail.

We use our benchmark, called \textsc{Dolomites} (short for
Domain-Specific Long-Form Methodical Tasks), to evaluate current
models in their ability to generate accurate and detailed outputs
(Section~\ref{sec:experiments}). We formulate the modeling problem as
long-form generation, where models are provided the task description
and the example input and asked to generate the example output. Our
experiments reveal that there is significant headroom in improving
performance on methodical tasks (Section~\ref{sec:results}) which are
inherently difficult (requiring reasoning skills and domain-knowledge),
and in terms of improving automatic evaluation of long-form text. In addition to well-known shortcomings
\citep{schluter-2017-limits,krishna-etal-2021-hurdles}, conventional metrics are not designed to capture expert knowledge.


We hope that
\textsc{Dolomites} can serve as a reference for domain-specific
use cases of language models and provide a means for evaluating future models. We
release our dataset and code at \url{https://dolomites-benchmark.github.io}.

\section{Problem Formulation}
\label{sec:problem}


In this section, we first describe the types of writing tasks
considered in this work. We refer to these tasks as methodical writing
tasks due to two properties that are common to their
execution. Firstly, each task requires \textbf{structured
  problem-solving}, where the task follows a specific order where each
step logically flows from the previous ones. For instance, in the \textit{Medicine}
task in Figure~\ref{fig:sample_tasks}, the task requires
producing an assessment of the patient, then a plan of care
and finally a full evaluation of the patient. Secondly, every task
usually follows a \textbf{consistent execution} across inputs, where
there is a standard specification of the input, the output and the
procedure for the task. In the same \textit{Medicine} task, given a patient's subjective and objective data,
the task structure and procedure would mostly stay consistent across patients.


To elicit descriptions of tasks from experts, we operationalize our
definition of a methodical task into a standard template (see
Figure~\ref{fig:sample_tasks} for examples). We require that every
task contains a brief \textit{task objective}, a \textit{task
  procedure} walking a beginner through how this task is conducted,
\textit{input and output sections}, which include information that is
typically given, and information that needs to be generated. Both
input and output sections are formatted in the form of section titles
and section descriptions. Finally, we collect \textit{additional
  notes} about the task, which can include best practices or common
mistakes, and missing context that is important when conducting the
task. 
%
%
%

We further expect our tasks to meet the following criteria: (1)~they are 
 \textbf{purely textual} and do  not involve other modalities in the
 input or output; (2)~they 
 \textbf{require domain expertise} and can only be  completed by an
 expert; (3)~they 
\textbf{do not require use of specific equipment or software}, with the
exception of searching the web; (4)~they are \textbf{frequent},
routinely performed by an expert at least once every few months; and
(5)~\textbf{time-consuming},  taking a significant but not
indefinite amount of time to complete (e.g.,  from a half hour to a
few days, but not  several months).


Aside from task descriptions, our dataset contains specific
\textbf{instantiations} of methodical tasks (see
Figure~\ref{fig:overview}). We create examples by populating
descriptions like those shown in Figure~\ref{fig:sample_tasks} with
plausible input and output sections (see
Section~\ref{sec:example_collection}).


\section{\textsc{Dolomites}: Data Curation}
\label{sec:data}

\setlength{\tabcolsep}{6pt} 

\renewcommand{\arraystretch}{1.15} 
\definecolor{Lavender}{rgb}{0.9,0.9,0.98}

\begin{table*}[t!]
\centering
\scalebox{.75}{
\rowcolors{2}{Lavender!55}{white}
\begin{tabular}{l@{~~}p{17cm}}
\rowcolor{Lavender!80}
\multicolumn{1}{c}{\textbf{Field}} & \multicolumn{1}{c}{\textbf{Sample Task Objective}} \\ \toprule
 Anthropology (8) & \textit{A survey to examine specific cultural practices, rituals, and societal norms within a cultural group or community} \\
 Architecture (20) & \textit{Developing a construction phasing plan for a building project}  \\
 Biology (21) & \textit{Developing a protocol for a toxicity assay}  \\
 Business (26) & \textit{Write a section of a non-financial report for a client, focusing on a company's environmental and social activities}  \\
 Chemistry  (21) & \textit{To write a retrosynthesis scheme/plan for a specific target molecule} \\
 Economics  (17) & \textit{Reviewing investment options for advising companies} \\
 Education  (23) & \textit{To create a lesson plan for a school class} \\
 Engineering (22) & \textit{To write the instructions for conducting a radioactive experiment.} \\
 Environmental Sci (23) & \textit{Writing the life cycle assessment of a system, product or process} \\
 Geography (20) & \textit{Analyzing the environmental and social impacts of illegal mining activities in a specific region} \\
 History (22) & \textit{Summarize and analyze a specific medieval legal code} \\
 Hospitality  (21) & \textit{Adapt existing recipes to cater to various dietary preferences} \\
 Journalism  (20) & \textit{Write a news story based on an interview} \\
 Law (38) & \textit{Drafting a petition to challenge a decision} \\
 Linguistics (20) & \textit{Carry out a short literature review of a given problem in linguistics} \\
 Literature (20) & \textit{To write a research proposal for a presentation at a literary research conference} \\
 Mathematics (15) & \textit{Writing an experimental setup suitable for testing a research hypothesis in applied mathematics} \\
 Medicine (24) & \textit{Writing a list of potential radiotherapy regimens for a cancer patient} \\
 Music (23) & \textit{Writing lyrics for a game's soundtrack} \\
 Philosophy (13) &  \textit{Provide ethical recommendations for patient/doctor cases} \\
 Physics  (21) & \textit{Design specifications for a pump or turbine system}  \\
 Political Sci (20) & \textit{Redline a management measure / legislative policy}  \\
 Psychology  (21) & \textit{Writing a study protocol of a neuroimaging research project} \\
 Sociology (20) & \textit{Analyzing responses from sociological interviews to identify themes relevant to the research question} \\
 Visual Arts (20) & \textit{The objective of this task is to write a catalog entry for an art exhibition} \\
 \bottomrule
\end{tabular}
}
\caption{Fields represented in \textsc{Dolomites}, with  number of tasks in parentheses and a sample task from each field.}
\label{tab:tasks_all}
\end{table*}

\subsection{Task Collection} 
\label{sec:task_collection}

In our data curation process, we first collect a typology of realistic
tasks that span multiple fields. These tasks are not meant to be
exhaustive, but instead represent realistic use
cases across fields.


\paragraph{Participants.} We recruit 266 participants 
from the Prolific crowdsourcing platform. We recruit
experts from 25 different fields, shown in Table~\ref{tab:tasks_all},
aiming for a broad coverage across disciplines. Participants qualify
as experts if they have formal education in the field, and at least 3
years of work experience. Additional details about the participants'
backgrounds are provided in Appendix~\ref{app:annotation}. 

\paragraph{Annotation Task.} We ask each annotator to provide
descriptions of two writing tasks they routinely perform in their
profession subject to the criteria listed in
Section~\ref{sec:problem}. For each task, annotators are asked to
fill in predefined fields (task objective, procedure, input and output
sections, additional notes), the same way as shown in
Figure~\ref{fig:sample_tasks}.
We ask annotators to give thorough descriptions as if they are
teaching a novice how to perform each task.
Instructions and interface screenshots are in
Appendix~\ref{app:annotation}.



\subsection{Task Analysis}

After collecting the initial set of tasks, we filter them manually
to ensure they meet the criteria outlined in
Section~\ref{sec:problem}, and obtain a total of 519~tasks. We find
that there are very few tasks from a field that are highly
similar.

Table~\ref{tab:tasks_all} provides the number of tasks across fields
and the task objective of a sample task from each field. Most fields
have at least 20 tasks, with some exceptions, where we were not able
to recruit as many experts. Across tasks, there are an average of $\sim$2.78 sections in the input and $\sim$2.82 sections in the output.
Collectively, tasks in \textsc{Dolomites} are cognitively demanding
and versatile in the types of reasoning they require. For instance, a
diagnostic task in medicine requires \emph{inductive reasoning} to go from
particular symptoms to a general diagnosis. Whereas, in legal
analysis, \emph{deductive reasoning} is required to reason about how laws are
interpreted in a specific case and \emph{analogical reasoning} is needed as
lawyers compare current cases with precedents. Similarly, in software
application design, \emph{abstract} reasoning is important while creativity is necessary for certain
tasks in the visual arts. While it is hard to describe
the type of reasoning required for all tasks,  every methodical task
essentially involves \textbf{analyzing} the input, 
\textbf{making inferences} based on the input and domain-specific knowledge,
and finally, \textbf{providing a justification} in writing.

\begin{figure*}[t!]
    \centering
    \includegraphics[width=2\columnwidth,keepaspectratio]{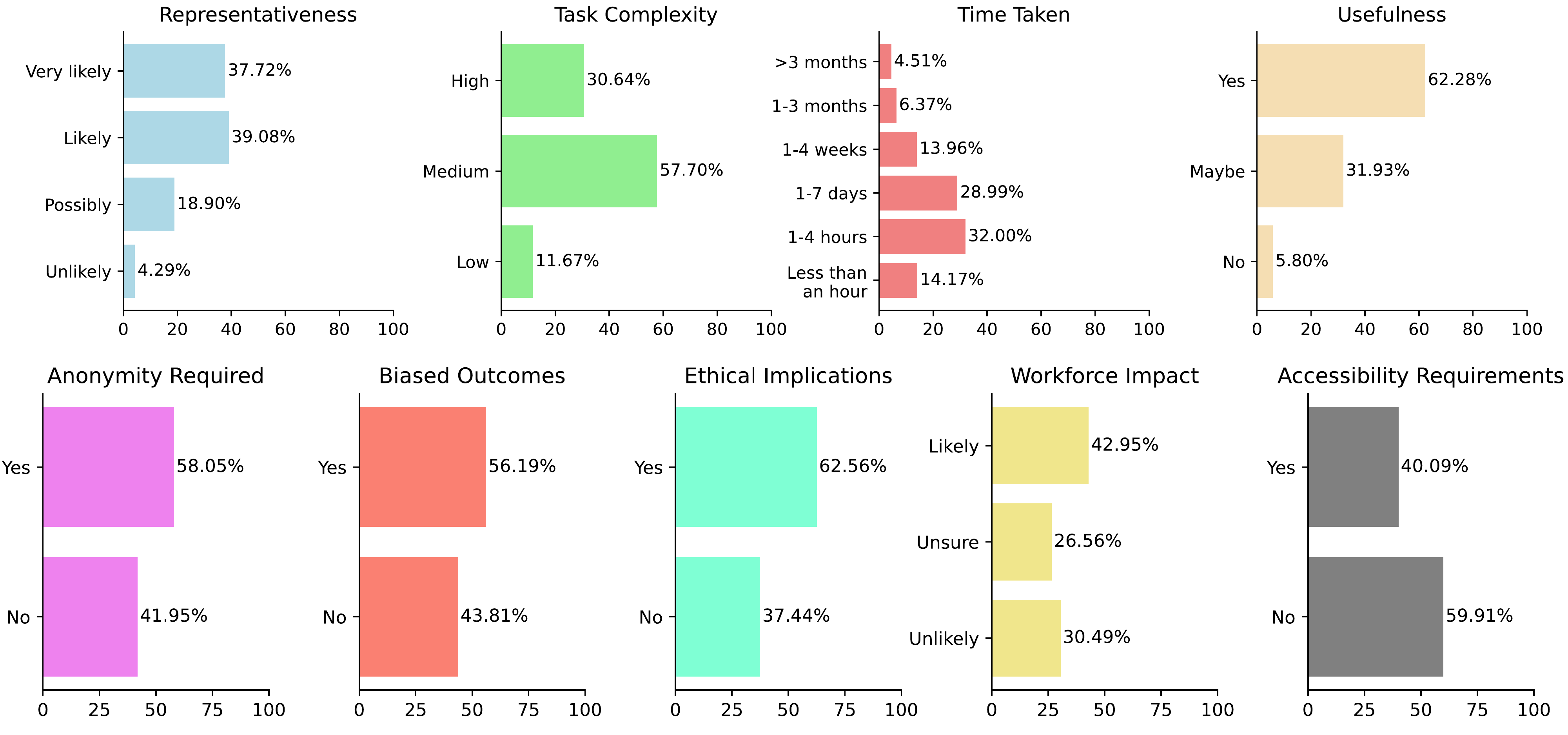}
    \vspace{-12pt}
    \caption{We conducted validation of methodical tasks in the \textsc{Dolomites} task collection by consulting an independent group of 3 experts from the field to which the task belongs. Here we show the Likert distributions of their ratings across various axes of importance. The question associated with each axis is listed in section~\ref{sec:task_validation}.}
    \label{fig:task_validation}
\end{figure*}

\subsection{Task Validation and Societal Implications}
\label{sec:task_validation}

We validate our collection of tasks by consulting an \emph{independent} group
of experts. 
Specifically, we collect Likert ratings for each task from three
experts on the following axes (the precise description for each item on
the scale is provided in Appendix~\ref{app:annotation}):

\begin{itemize}[noitemsep, leftmargin=*, label=\small{$\bullet$}]
 \item \textbf{Representativeness}: How likely is this task to be conducted by an expert in your field?
    \item \textbf{Complexity}: How would you rate the complexity of this task?
    \item \textbf{Time Required}: How much time is typically required to complete this task?
\item \textbf{Usefulness}: Would you or other experts find it useful
  if an AI system were to propose initial outputs for this task (which may be lacking), that can be validated and improved by experts?
\end{itemize}

The question above is motivated by prior work showing that AI writing
assistants could significantly benefit the productivity of experts
\citep{eloundou2023gpts,noy2023experimental,dell2023navigating}. Beyond
productivity, we were additionally interested in expert opinions about
the \emph{societal implications} of using language models as writing
assistants. Hence, for each task, we elicit answers to the following
questions which require a free-text response in addition to a Likert
rating.
\begin{itemize}[noitemsep, leftmargin=*, label=\small{$\bullet$}]
\item \textbf{Anonymity Required}: Is it important to ensure anonymity of any individuals or organizations if an AI system is used for conducting this task? 
\item \textbf{Biased Outcomes}: Could relying on automatically generated outputs for this task result in biased or potentially harmful decisions for certain groups of people?
\item \textbf{Ethical Considerations}: Are there ethical
  considerations (e.g., privacy, copyright issues) associated with the use of AI systems for this task? 
\item \textbf{Workforce Impact}: Could partial automation of this task have an impact on the workforce in the short term?
\item \textbf{Accessibility Requirements}: Would the use of AI tools
  for this task require making exceptional considerations to ensure accessibility?
\end{itemize}


The main outcomes of our validation study are presented in
Figure~\ref{fig:task_validation}. We find that the tasks collected in
\textsc{Dolomites} are ecologically valid, i.e.,~they are likely
($\sim$76\%) to be conducted by field experts. Most of them are of
medium or high complexity, requiring moderate (few years of
experience) or substantial (several years of experience)
expertise. While they are complex tasks, judgements about time taken
reveal that most ($\sim$61\%) would take an expert from 1 hour to 7
days to complete. This degree of difficulty suggests that it is
conceivable for language models to be useful assistants for these
tasks. Finally, an overwhelming majority of experts would be
interested ($\sim$62\%) or open to trying ($\sim$32\%) to use a
language model that proposes initial outputs for the task.



With regard to the societal implications of language model use, the
need for anonymity emerged as a concern for a significant number of
tasks ($\sim$58\%). In fields like medicine, psychology,
and law, experts emphasized the importance of protecting
patient/client confidentiality.
Similarly, experts felt strongly that 
proprietary information and trade secrets should be kept private in
fields like business.
 Experts further thought that using language model responses without
 careful perusal can result in biased outcomes ($\sim$56\% of tasks)
 which could affect marginalized or underrepresented groups. They also
 raised various ethical concerns relating to copyright issues, privacy
 issues and stifling of human creativity due to over-reliance on AI.

Many experts ($\sim$43\%) recognized that partial automation of
writing tasks is likely to impact the workforce in the short term,
potentially leading to changes in job roles or skill requirements. At
the same time, they were optimistic that this would improve
productivity and bring positive changes to the nature of the work.
It is important to ensure that users of all backgrounds and
capabilities have equal access to language models.
A significant number of tasks ($\sim40\%$) were rated as requiring
exceptional considerations to be made for ensuring accessibility to
all users.
%
%
%
Across fields, experts highlighted that while language models as
writing assistants can improve productivity, human oversight is
important for responsible use of these technologies.

\begin{figure*}[t!]
    \centering
    \includegraphics[width=2\columnwidth]{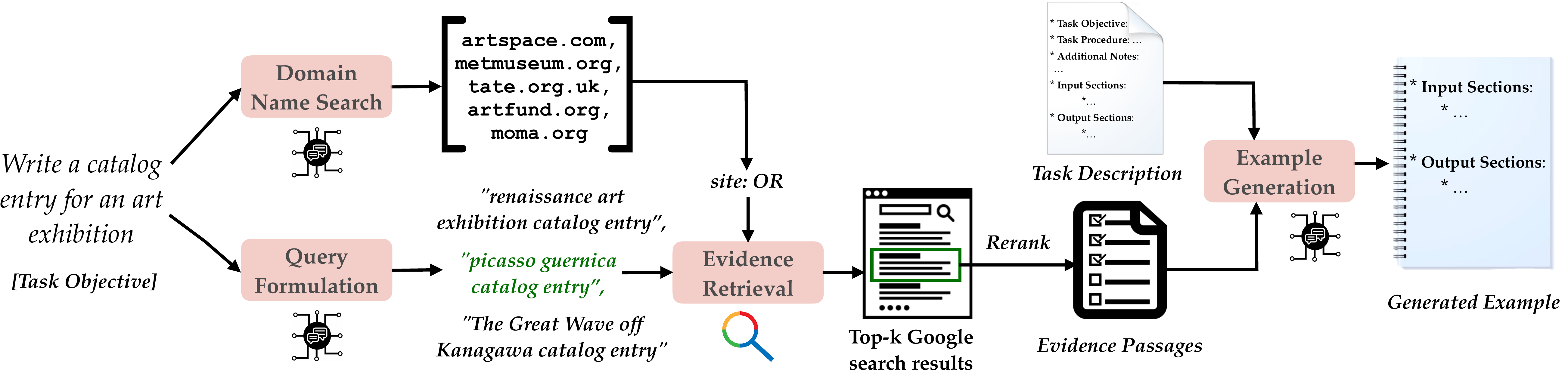}
    \caption{Here, we outline the method for constructing examples of tasks in \textsc{Dolomites}. Using the task objective for a task, we first generate more specific queries to search for relevant web documents, where we constrain our search to authoritative domain names for the task. Using a set of retrieved evidence passages and the complete task description, we then generate an example of the task that fits the task structure using a language model. This example is then post-edited by the same expert who provided the task (further described in section~\ref{subsubsec:post-editing}).}
    \vspace{-10pt}
    \label{fig:example_creation}
\end{figure*}

\subsection{Example Collection}
\label{sec:example_collection}

To evaluate language model capabilities in assisting experts with
their tasks, we create examples of input and output sections
with concrete details. We adopt a human-in-the-loop methodology, where
initial examples are generated by a model, which are then
post-edited by the same expert who provided the task. We describe this
process below.

\subsubsection{Retrieval-Augmented Generation}

We believe that samples of methodical tasks are partially available in
documents on the web. Hence, we retrieve relevant documents for each
task and generate examples by prompting models with passages from
these documents as context. This process is depicted in
Figure~\ref{fig:example_creation} and explained below.

\paragraph{Query Formulation.}  Given a task description, we first
need to find more specific queries that could potentially result in
relevant web documents. For example, for  writing
a catalog entry for an art exhibition, search queries like  ``\textsl{renaissance art exhibition catalog
  entry}'' or  ``\textsl{picasso guernica catalog entry}''
are likely to result in documents that contain examples of the
task. To generate search queries, we prompt Bard
\citep{manyika2023overview} (with 1 exemplar) with the task objective
and instruct it to generate more specific queries that can help find
web documents which contain demonstrations of the task
(Table~\ref{tab:query_formulation_prompt} in
Appendix~\ref{app:experiments}). We generate 10 search queries for each
task and restrict search to reliable and authoritative
sources. These are collected by prompting Bard (with 1~exemplar) with
the task objective to generate URLs to domain names which will be useful to
find real examples for the given task
(Table~\ref{tab:domain_name_prompt} in
Appendix~\ref{app:experiments}).

\paragraph{Evidence Collection.}  Using each search query, we
gather the top-10 documents from  Google search restricted to
relevant domains with the \texttt{site:} operator in the query.
Documents are then split into passages of
4,000~characters with a 100 character sliding window.

\paragraph{Conditional Generation.} Having gathered evidence
which may contain task demonstrations, we generate
examples by prompting models with this evidence. We explore a
\emph{multi-document} setting, where passages are sampled from
multiple documents and a \emph{single document setting}, where
passages are sampled from a single document as we found that the
appropriate choice depends on the task.\footnote{For instance, a
task that requires drafting a legal opinion might benefit from
multiple relevant documents whereas a single document might be
sufficient for writing the catalog entry for an artwork.}
Passages are reranked using an in-house reranker and the top-5
passages are provided as context, along with the task
description, to a large language model (Gemini-Ultra
\citep{team2023gemini} and Bard in our case), which is asked to
generate an example of the task
(Table~\ref{tab:example_generation_prompt} in Appendix~\ref{app:experiments}). Not all information
mentioned in the task description is required to be present in
the context and the model is allowed to infill content
to construct an example. We generate up to 10~examples for each
task.

\subsubsection{Expert Post-Editing}
\label{subsubsec:post-editing}
\begin{figure*}[ht!]
    \centering
    \includegraphics[width=\textwidth]{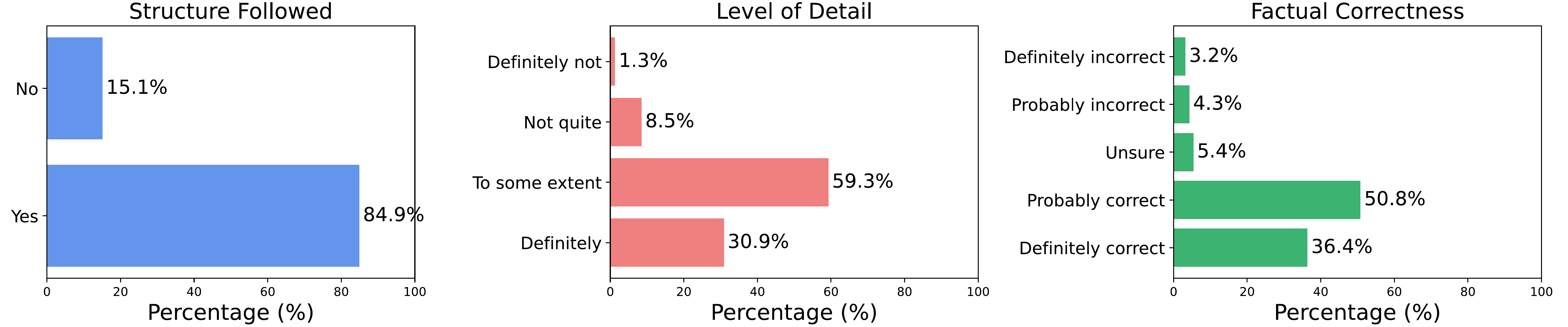}
    \caption{Expert judgements of  original examples along three
      dimensions: \textbf{task structure followed} (whether the example includes all the input and output sections from the task description), \textbf{level of detail} (whether the example shows a detailed and concrete sample of the task) and \textbf{factual correctness} (whether the example is factually correct).}
    \label{fig:editing_qa}
\end{figure*}

Even though they are based on retrieved web documents, model-generated
examples could have many issues. They may not adhere
to the structure specified in the task description, they may contain
factual inconsistencies or inaccuracies, lack in depth, or be vague.
To remedy these issues, we present the examples to the same
experts who wrote the tasks for post-editing. They are asked to choose
the most plausible example out of four variants
, and use single or multi-document evidence.

Prior to post-editing, experts are asked to label examples according
to three criteria on a Likert scale: adherence to task structure,
factual correctness, and level of detail. They are also shown 1)~the
evidence passages for the example and 2)~a critique (generated using
Gemini-Ultra with the prompt in Appendix~\ref{app:experiments},
Table~\ref{tab:critique_prompt}) that may not be comprehensive, to aid
them in identifying issues with the example. The critique is provided
to make post-editing more efficient. They are required to fix any
valid issues they recognize in the example as well as any valid issues
identified by the critique.

\subsubsection{Example Quality Analysis}

\begin{figure}[t!]
  \vspace{-1.06cm}
  \begin{tabular}{@{\hspace*{-.6ex}}l@{~}l@{}}
           \raisebox{-.8cm}[0pt]{{\small a.}} & \raisebox{-.8cm}[0pt]{{\small b.}} \\
    \includegraphics[width=.54\columnwidth,keepaspectratio]{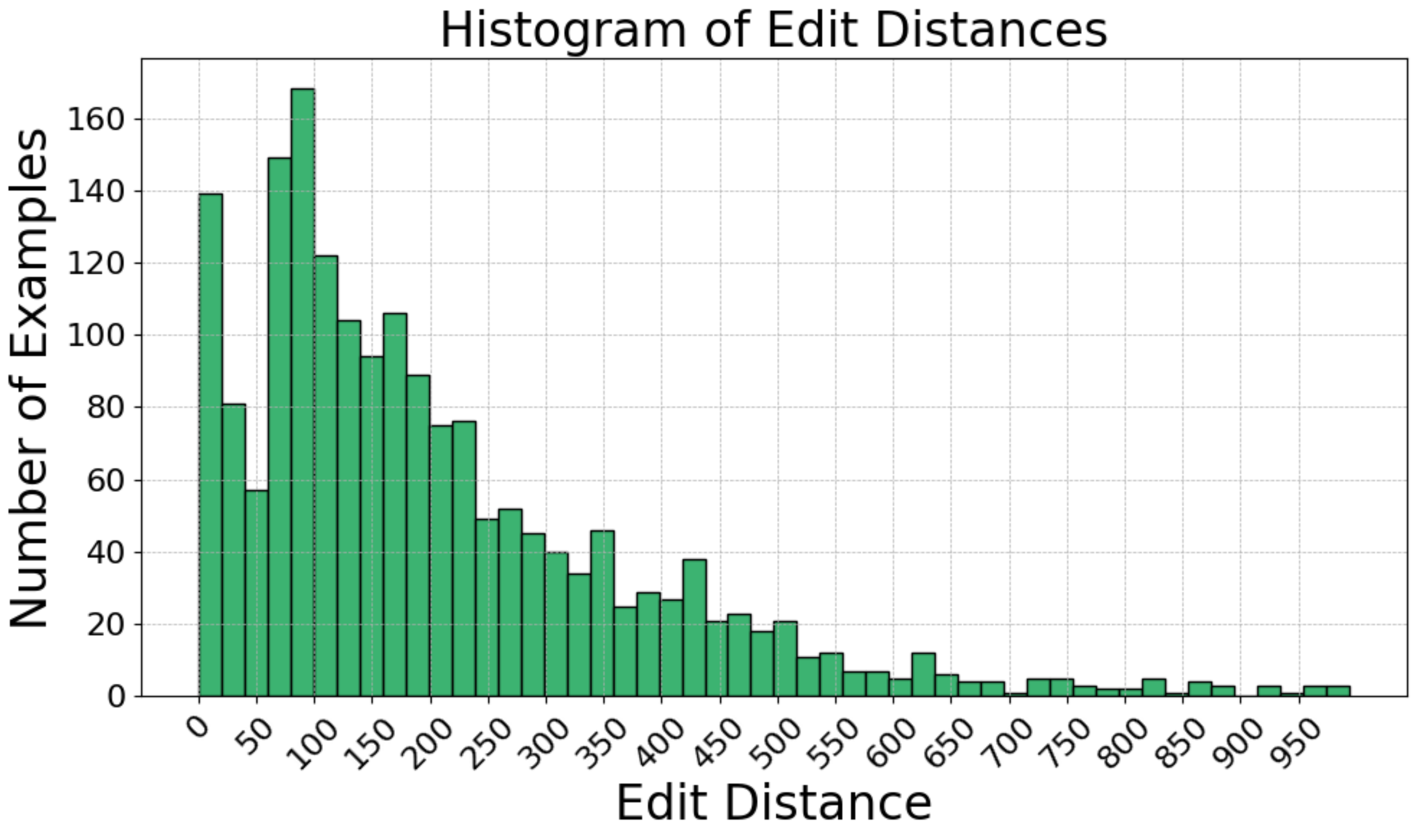}
& \includegraphics[width=.5\columnwidth]{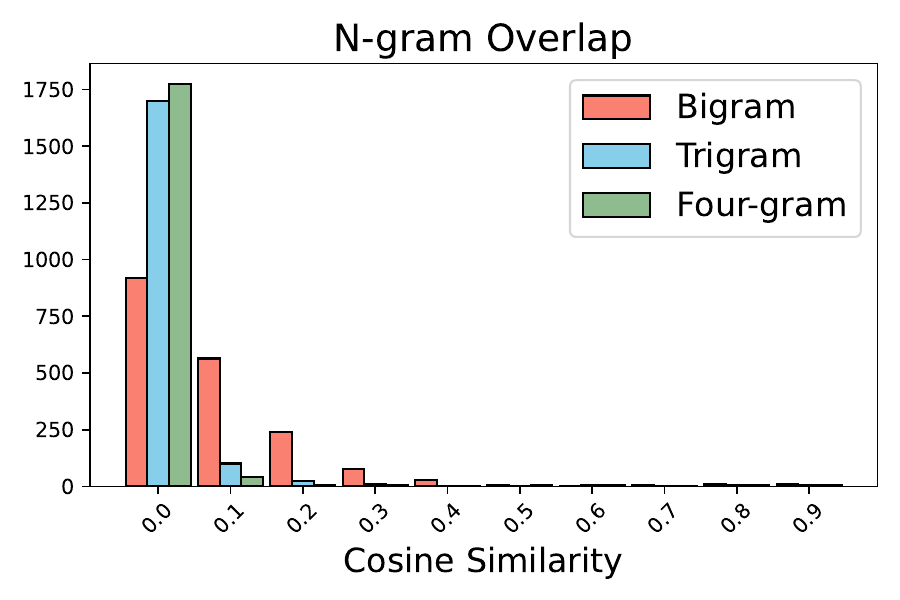}        \\
  \end{tabular}
  \caption{\label{fig:edit_dists} Histogram of (a)~word-level edit distance between 
original and post-edited example and (b)~cosine similarities based on n-grams in  post-edited examples and evidence used as context.}
  \vspace{-10pt}
\end{figure}

\paragraph{Automatic Analysis.} Expert judgements of automatically
generated examples are shown in Figure~\ref{fig:editing_qa}. We find
that the majority of examples already follow the task structure
($\sim$85\%)  and most of them are
\textit{probably} or \textit{definitely} correct. However, a large
number of examples are lacking in depth and detail. 
We show a histogram of the word-level edit distance between all tokens in the
original and edited example in Figure~\ref{fig:edit_dists}a and
relevant statistics of the original and post-edited examples in
Table~\ref{tab:edit_stats}. The histogram suggests that on average,
there are significant changes made to the original examples during
post-editing. Since most experts judge that examples are lacking in
depth, the edited examples are expectedly much longer on average. The
edited examples also adhere better to the task description (on
average, 97.67\% sections in the task description are found in the
edited examples compared to 92.81\% in the original
examples). We analyzed a random set of 100 examples and labeled each example with the type(s) of edits it contained. We found broadly the following types of edits: fact addition (88\%), fact deletion (20\%), fact update (65\%), stylistic rewrites (76\%) and reorganization (23\%). These edit types are described further in Appendix~\ref{app:annotation}. Finally, we compute readability scores of the examples as a
noisy approximation of the complexity and level of detail in the text,
using the Flesch-Kincaid Grade Level test
\citep{kincaid1975derivation}. Higher readability values indicate that
a piece of text requires more formal education and expertise to
understand. We find that post-edited examples have higher scores,
possibly indicating higher level of technical depth.





\renewcommand{\arraystretch}{1} 

\begin{table}[t!]
\centering
\scalebox{.7}{
\begin{tabular}{ lccccc }
\rowcolor{gray!40}
 & \textbf{Avg Length} & \textbf{\makecell{Avg section\\presence \%($\uparrow$)}} & \textbf{\makecell{Flesch-Kincaid\\Grade Level ($\uparrow$)}} \\ \toprule
\texttt{original} & 388.98 & 92.81 & 11.69 \\
\texttt{post-edited} & 590.24 & 97.67 & 13.46  \\
 \bottomrule
\end{tabular}
}
\caption{Statistics of the original and post-edited examples in \textsc{Dolomites}.}
\vspace{-15pt}
\label{tab:edit_stats}
\end{table}

\begin{figure*}[ht!]
    \centering
    \includegraphics[width=\textwidth,keepaspectratio=True]{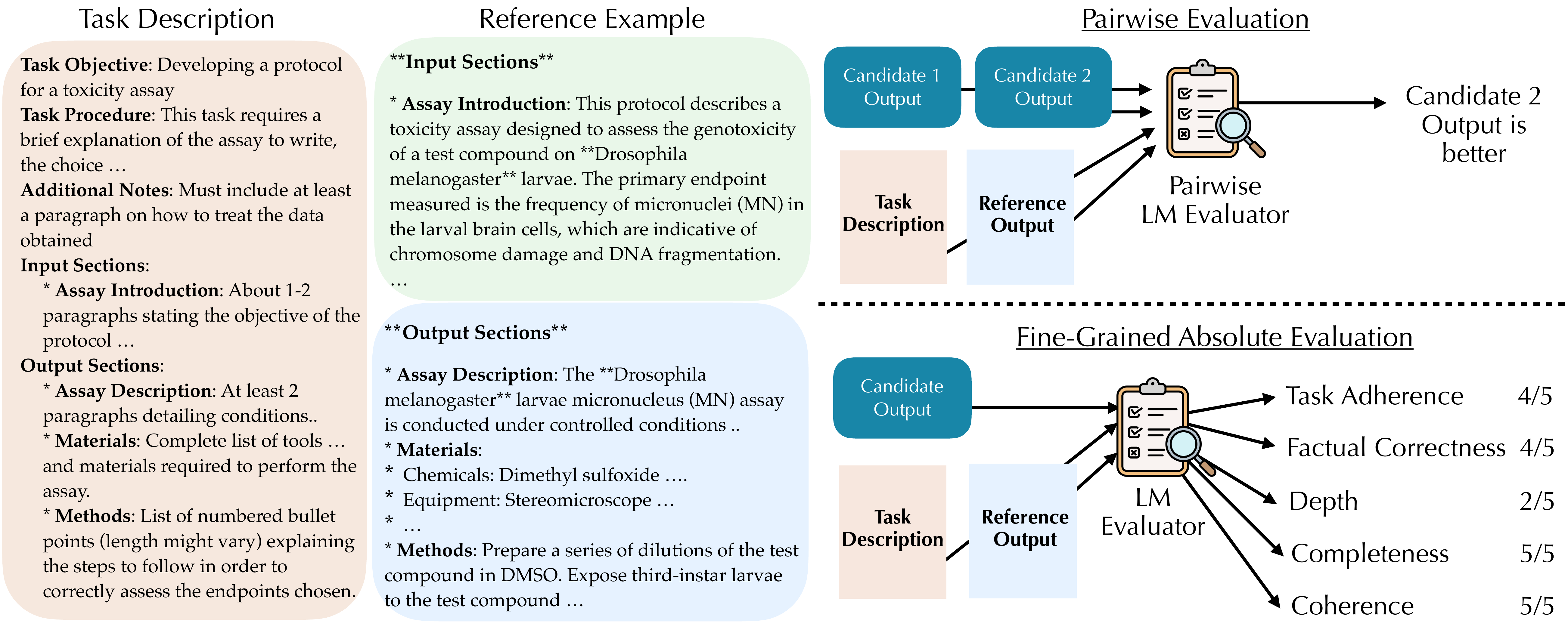}
    \caption{Automatic evaluation protocol for \textsc{Dolomites}, involving two modes of evaluation: \textbf{pairwise evaluation} of two candidate outputs and \textbf{fine-grained absolute evaluation} of a candidate output on our proposed axes.}
    \label{fig:evaluation}
    \vspace{-10pt}
\end{figure*}



\paragraph{Data Contamination Analysis.} Examples in \textsc{Dolomites} are created by conditioning on passages from web documents. We do
not require these documents to contain complete examples of the task,
and models are allowed to infill information to create an
example. However, if these documents are seen by models during
large-scale pretraining, it might be difficult to conduct a clean
evaluation. We examine whether this is the case by computing the
similarity of the post-edited examples to the evidence passages
provided during generation. Figure~\ref{fig:edit_dists}b plots the
cosine similarity between the n-grams found in the post-edited
examples and evidence used as context.  As can be seen, similarity to
retrieved passages is fairly low in most examples, which suggests a
low risk of memorization due to pretraining.

In addition, we check directly if text in our examples can be found in large pretraining corpora. We use an open-source toolkit called WIMBD \cite{elazar2024whats} to measure the presence of text in generated examples in two large pretraining corpora: C4 \cite{raffel2020exploring} (364 million documents) and Dolma \cite{soldaini2024dolma} (5,245 million documents). These corpora are widely used to train large language models \cite{raffel2020exploring,dodge-etal-2021-documenting,chowdhery2023palm,groeneveld2024olmo}. We follow prior work from \citet{chowdhery2023palm} and determine an example as contaminated if 70\% of all possible 8-grams in the example were seen at least once in the pretraining corpus. Conducting this process for a random sample of 100 examples in \textsc{Dolomites}, we find that none of the examples are contaminated.



\section{Experiments}
\label{sec:experiments}

\renewcommand{\arraystretch}{1.1} 

\begin{table*}[!t]
\centering
\scalebox{.85}{
\setlength\tabcolsep{4pt} 
\begin{tabular}{ccccc p{2.5cm} p{2.5cm} c}
    \toprule
    \rowcolor{gray!40}
    \textbf{Model} & \textbf{\makecell{Avg\\Length}} & \textbf{\makecell{Avg Section\\Presence \%}} & \textbf{BLEURT} & \textbf{ROUGE-L} & \textbf{\makecell{nli \\ (forward)}} & \textbf{\makecell{nli \\ (reverse)}} & \textbf{\makecell{nli \\ (h-mean)}} \\
    \midrule
    \texttt{Claude-3 Opus} & 417.41 & 91.53 & 0.4156 & 0.2395 &  0.3584$_{[0.34,0.38]}$  & 0.3769$_{[0.36,0.40]}$ & 0.3674 \\
    \texttt{Command-R-Plus} & 440.44 & 92.92 & 0.4068 & 0.2134 & 0.3926$_{[0.37,0.41]}$ & 0.3623$_{[0.34,0.38]}$ & 0.3768 \\
    \texttt{Gemini-1.5-Pro} & 349.27 & 95.25 & 0.4136 & 0.2371 & \textbf{0.4065}$_{[0.39,0.42]}$  & 0.3846$_{[0.37,0.40]}$ & 0.3952 \\
    \texttt{Gemini-1.5-0409} & 361.87 & \textbf{95.74} & 0.4068 & 0.2361 & 0.3994$_{[0.38,0.42]}$ & 0.3984$_{[0.38,0.42]}$ & \textbf{0.3989} \\
    \texttt{Gemini-Pro} & 269.68 & 93.61 & 0.4124 & 0.2280 & 0.3415$_{[0.32,0.36]}$ & 0.3090$_{[0.29,0.33]}$ & 0.3244 \\
    \texttt{GPT-3.5-Turbo} & 240.89 & 88.98 & \textbf{0.4276} & 0.2309 & 0.3854$_{[0.37,0.40]}$ &  0.2949$_{[0.28,0.31]}$ & 0.3341 \\
    \texttt{GPT-4} & 407.35 & 95.46 & 0.4155 & 0.2271 & 0.3934$_{[0.38,0.41]}$ & \textbf{0.3993}$_{[0.38,0.42]}$  & 0.3963 \\
    \texttt{Mistral-Large} & 327.12 & 92.61 & 0.4158 & 0.2390 &  0.3524$_{[0.33,0.37]}$  &  0.3523$_{[0.33,0.37]}$  & 0.3523 \\
    \texttt{Mixtral-8x22B} & 339.16 & 95.16 & 0.4212 & \textbf{0.2450} &  0.3951$_{[0.38,0.41]}$  &  0.3583$_{[0.34,0.37]}$  & 0.3758 \\
    \texttt{Mixtral-8x7B} & 386.61 & 88.39 & 0.4098 & 0.2266 &  0.3290$_{[0.31,0.35]}$  &  0.3097$_{[0.29,0.33]}$  & 0.3191 \\
    \texttt{OLMo-7B-Instruct} & 784.22 & 74.02 & 0.3905 & 0.1752 &  0.1929$_{[0.18,0.21]}$  &  0.1721$_{[0.16,0.19]}$  & 0.1819 \\
    \bottomrule
\end{tabular}
}
\caption{Results on the \textsc{Dolomites} test set with standard metrics and factual consistency using NLI models. We report 95\% confidence intervals along with the average NLI scores.}
\label{tab:results_std}
\end{table*}

\subsection{Setup and Models}

We create a development-test split for \textsc{Dolomites}, with 820 examples
in the development (dev) set and 1,037 examples in the test set. There are
172~\emph{seen} tasks with examples  in both dev and test and
99 \emph{unseen} tasks with examples only in the test set.

For evaluation, we considered multiple performant models from
various companies as well as open-source models. In all cases, we favored instruction-tuned variants
because of their better performance on other benchmarks. Specifically,
we report experiments with Claude-3 Opus
\citep{TheC3}, Command-R-Plus \citep{gomez2024cmdrplus},
Gemini-1.5-Pro and Gemini-1.5-0409 \citep{reid2024gemini} and \mbox{Gemini-Pro}
\citep{team2023gemini}, GPT-3.5-Turbo and \mbox{GPT-4}
\citep{OpenAI2023GPT4TR}, Mixtral-8x7B and Mixtral-8x22B
\citep{jiang2024mixtral} and Mistral-Large \citep{mistral-large}, and
OLMo-7B-Instruct \citep{groeneveld2024olmo}.  In all cases, we prompt
models with the task description, the input sections corresponding to
an example, and instruct them to generate the output sections for the
example, in a zero-shot manner. Hyperparameters,
prompts and model identifiers are in Appendix~\ref{app:experiments}.

\subsection{Automatic Evaluation}
\label{subsec:automatic_evaluation}

\subsubsection{LM-based Evaluation}
\label{subsubsec:lm_eval}

\paragraph{LM-based Pairwise Evaluation.} Language models are being increasingly
used as evaluators \citep{chiang-lee-2023-large}, and our primary evaluation also involves using LMs as evaluators.
However, recent work points out that LM judgements can be misleading and biased \citep{shen-etal-2023-large,wang2023large,zheng2023judging,panickssery2024llm}.
We use \emph{multiple} language models as judges to give preferences
for a pair of model outputs. While this does not alleviate the problem
of biased LM judgements, we believe it is slightly more reliable since we are not biased by a single model's judgements. In
all comparisons, we use one of the strongest models, GPT-4, as the base
comparison model. We sample outputs on the test set from a candidate
model and GPT-4 (randomizing their order in the prompt) and ask the evaluator model to judge which output is
better and provide a justification.
We consider three models as evaluators: GPT-4,
Claude-3 Opus and Gemini-1.5-0409. The win rate is computed by summing up
the number of wins for a candidate model plus half the number of ties.

\paragraph{LM-based Fine-Grained Evaluation.} In addition to preference judgements, we evaluate models on five finer grained aspects of response quality: task adherence, factual correctness, depth, completeness and coherence\footnote{These aspects of response quality are defined in Table \ref{tab:lm_eval_prompt_fine}.}. We use GPT-4 to collect absolute ratings (on a scale of 1-5) of individual model responses on each of these aspects.

\subsubsection{Other Evaluation Metrics}
\paragraph{Round-Trip Factual Consistency.} We also 
measure the extent to which statements in the model output are
consistent with statements in the reference output. We compute 1) \emph{forward}
entailment considering a reference section as the premise and the
corresponding model section as the hypothesis and 2) \emph{reverse}
entailment considering a model output section as the premise and the
corresponding reference section as the hypothesis. Scores are
aggregated over all sections and examples. These metrics loosely
capture the notions of precision and recall, we also report the
harmonic mean of the two. We use the TRUE model
\cite{honovich-etal-2022-true-evaluating} to predict entailment scores
(ranging from 0 to 1) and report 95\% confidence intervals. Note that this metric has weaknesses, as it assumes that there is a single valid reference for each example, which may not be true for many examples in \textsc{Dolomites}.

\paragraph{Conventional Metrics.} Prior work has recognized that
conventional metrics for text generation are lacking in various ways
\cite{liu-etal-2016-evaluate,novikova-etal-2017-need,krishna-etal-2021-hurdles}. Nevertheless,
for completeness, we report results with ROUGE-L \cite{lin-2004-rouge}
and BLEURT \cite{sellam-etal-2020-bleurt}. In addition, we report the
average output length and average section presence (i.e., the
percentage of output sections specified in the task that are present in the
generated output, averaged across all examples) as a measure of
instruction-following capabilities. Note that the average length of
reference outputs is 341.42~tokens.

\section{Results and Discussion}
\label{sec:results}

\begin{figure}[ht!]
    \centering
    \includegraphics[width=.5\textwidth]{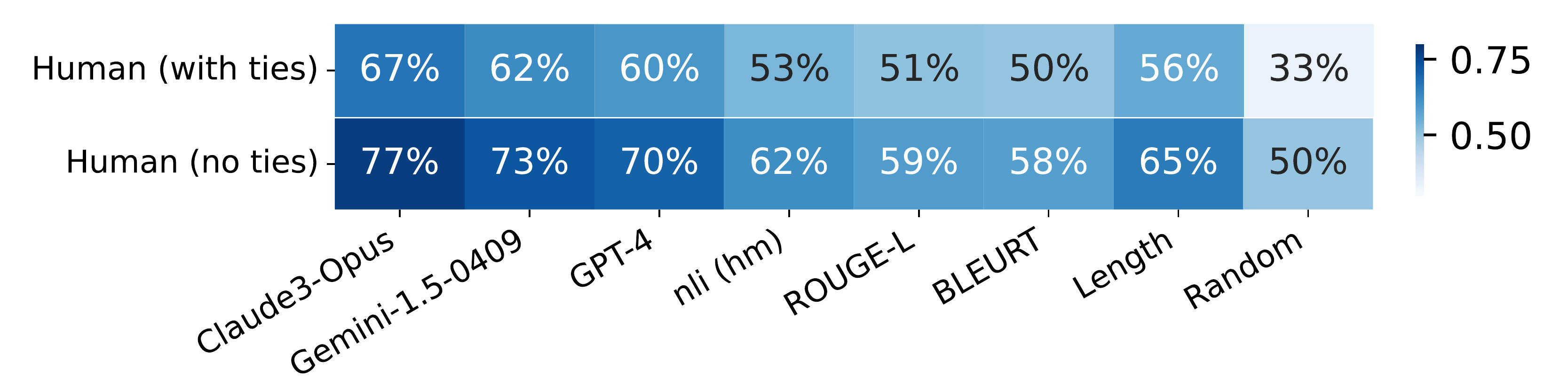}
    \vspace{-15pt}
    \caption{Percentage agreement of automatic evaluation measures with human labels. Pairwise judgements from Claude-3 Opus have the highest correlation with human labels.}
    \vspace{-10pt}
    \label{fig:agreement}
\end{figure}

\paragraph{Human Evaluation.} To evaluate the reliability of automatic metrics, we measure how well the above automatic evaluation measures correlate with human judgements. Specifically, we sample 200 pairs of model outputs, where each pair comes from two randomly chosen models. We (the authors) then label which model output is better (or if they are tied) according to their task adherence, factual correctness and depth\footnote{Human agreement between two annotators was found to be 75\% on 100 examples while including ties.}. On these 200 pairs, we also get automatic preference judgements from all the evaluation measures discussed in Section~\ref{subsec:automatic_evaluation} (we convert float scores for two outputs into binary judgements). The percentage agreements between human labels (with and without pairs with ties) and all evaluation measures are in Figure~\ref{fig:agreement}. In summary, we find that \textbf{LM-based evaluation measures have the highest correlations with human judgements}, followed by the NLI measures and then ROUGE-L and BLEURT. We note that an evaluator that always picks the longer output also has reasonable agreement rates.

\begin{table}[t]
\centering
\scalebox{.65}{
   \setlength\extrarowheight{-4pt}
\begin{tabular}{lccc}
\rowcolor{gray!40}
    \toprule
    \textbf{Model} & \multicolumn{1}{c}{\textbf{GPT-4}} & \multicolumn{1}{c}{\textbf{Claude-3 Opus}} & \multicolumn{1}{c}{\textbf{Gemini-1.5-0409}} \\
    \midrule
    \texttt{Claude-3 Opus} &  48.1 & 52.7 & 49.6 \\
    \texttt{Command-R-Plus} & 34.4 & 45.8 & 38.7 \\
    \texttt{Gemini-1.5-Pro} & 41.1 & 46.1 & 51.0 \\
    \texttt{Gemini-1.5-0409} & 42.9 & \textbf{55.4} & \textbf{60.9} \\
    \texttt{Gemini-Pro} & 17.6 & 21.0 & 22.1 \\
    \texttt{GPT-3.5-Turbo} & 12.2 & 11.5 & 12.4 \\
    \texttt{GPT-4} & \textbf{50.0} & 50.0 & 50.0 \\
    \texttt{Mistral-Large} & 27.2 & 28.8 & 26.7 \\
    \texttt{Mixtral-8x22B} & 21.6 & 25.1 & 17.6 \\
    \texttt{Mixtral-8x7B} & 17.8 & 23.5 & 15.6 \\
    \texttt{OLMo-7B-Instruct} & \hspace{.18cm}4.2 &  \hspace{.18cm}5.5 & \hspace{.18cm}3.3 \\
    \bottomrule
\end{tabular}
}
\caption{Model win rates ($\pm3$) against GPT-4 on the \textsc{Dolomites} test
  set using three LM-based autoraters (GPT-4, Claude-3 Opus, and
  Gemini-1.5-PP). GPT-4's win rate is 50\% since it is the base comparison model.
  Note that pairwise judgements from Claude-3 Opus have the highest correlation with human judgements.}
  \vspace{-15pt}
\label{tab:results_lm}
\end{table}

\renewcommand{\arraystretch}{1.1} 

\begin{table*}[!ht]
\centering
\scalebox{.75}{
\setlength\tabcolsep{12pt} 
\begin{tabular}{lccccc:c}
    \toprule
    \rowcolor{gray!40}
    \textbf{Model} & \textbf{\makecell{Task\\Adherence}} & \textbf{\makecell{Factual \\Correctness}} & \textbf{Depth} & \textbf{Completeness} & \textbf{Coherence}  & \textbf{\makecell{Average Rating}} \\
    \midrule
    \texttt{Claude-3 Opus} & 4.57 & 4.73 & 4.36 & 4.54 & 4.84 & \textbf{4.61} \\
    \texttt{Command-R-Plus} & 4.36 & 4.57 & 4.17 & 4.35 & 4.73 & 4.44 \\
    \texttt{Gemini-1.5-Pro} & 4.41 & 4.70 & 4.21 & 4.38 & 4.82 & 4.50 \\
    \texttt{Gemini-1.5-0409} & 4.49 & 4.71 & 4.30 & 4.46 & 4.83 & 4.56 \\
    \texttt{Gemini-Pro} & 3.95 & 4.24 & 3.62 & 3.87 & 4.43 & 4.02 \\
    \texttt{GPT-3.5-Turbo} & 3.90 & 4.34 & 3.37 & 3.73 & 4.36 & 3.94 \\
    \texttt{Mistral-Large} & 4.38 & 4.59 & 3.98 & 4.31 & 4.72 & 4.40 \\
    \texttt{Mixtral-8x22B} & 4.23 & 4.47 & 3.83 & 4.18 & 4.60 & 4.26 \\
    \texttt{Mixtral-8x7B} & 3.99 & 4.23 & 3.64 & 3.94 & 4.36 & 4.03 \\
    \texttt{OLMo-7B-Instruct} & 2.59 & 3.04 & 2.62 & 2.61 & 2.95 & 2.76 \\
    \bottomrule
\end{tabular}
}
\caption{Results on the \textsc{Dolomites} test set along fine-grained aspects of response quality. All ratings are performed on a scale of 1-5 by GPT-4-Turbo-Preview and we report average ratings across all examples (all average ratings were found to be statistically different from the best model's average ratings, which is in this case, Claude-3 Opus).}
\vspace{-10pt}
\label{tab:results_fine}
\end{table*}

We first report overall statistics of model responses and scores using conventional metrics in
Table~\ref{tab:results_std}. Based on the average section presence, we
note that models are largely effective at generating almost all
relevant output sections from the task description. A few models
(GPT-4 and Gemini-1.5-0409) excel more at following this instruction. Based
on the NLI scores, we note that the nli (reverse) scores are on
average lower than nli (forward), which suggests that generated
outputs contain statements not entailed by the reference,
e.g.,~because they are inaccurate or irrelevant. We observe that
Gemini-1.5-0409 and GPT-4 produce more information that is factually
consistent with the reference, while Claude-3 and
Command-R-Plus are also performant.
Once again, we note that reference-based metrics generally penalize outputs which are valid but different from the reference, and there might not just be a single reference for some of these examples, especially when the task is more subjective.

\paragraph{Pairwise LM Evaluation Results.} We show the win rates according to different LM evaluators in Table~\ref{tab:results_lm}. Based on these win rates, we note that a few models such as Claude-3 Opus, Gemini-1.5-Pro and Gemini-1.5-0409 prove to be comparable to GPT-4. We also report win rates with a length penalty for longer outputs in Table~\ref{tab:results_lm_lp} and the overall rankings do not change.

\paragraph{Fine-Grained LM Evaluation Results.} Finally, we show ratings of models according to finer-grained aspects of response quality in Table~\ref{tab:results_fine}. The overall conclusions are roughly similar, i.e., Gemini-1.5-0409, Claude-3 Opus and GPT-4 have the highest ratings across axes. Across the axes considered in our rubric, models struggle most with the level of technical depth of the generated text.

\begin{figure}[t!]
    \centering
    \hspace*{-.2cm}\includegraphics[width=0.5\textwidth,keepaspectratio=True]{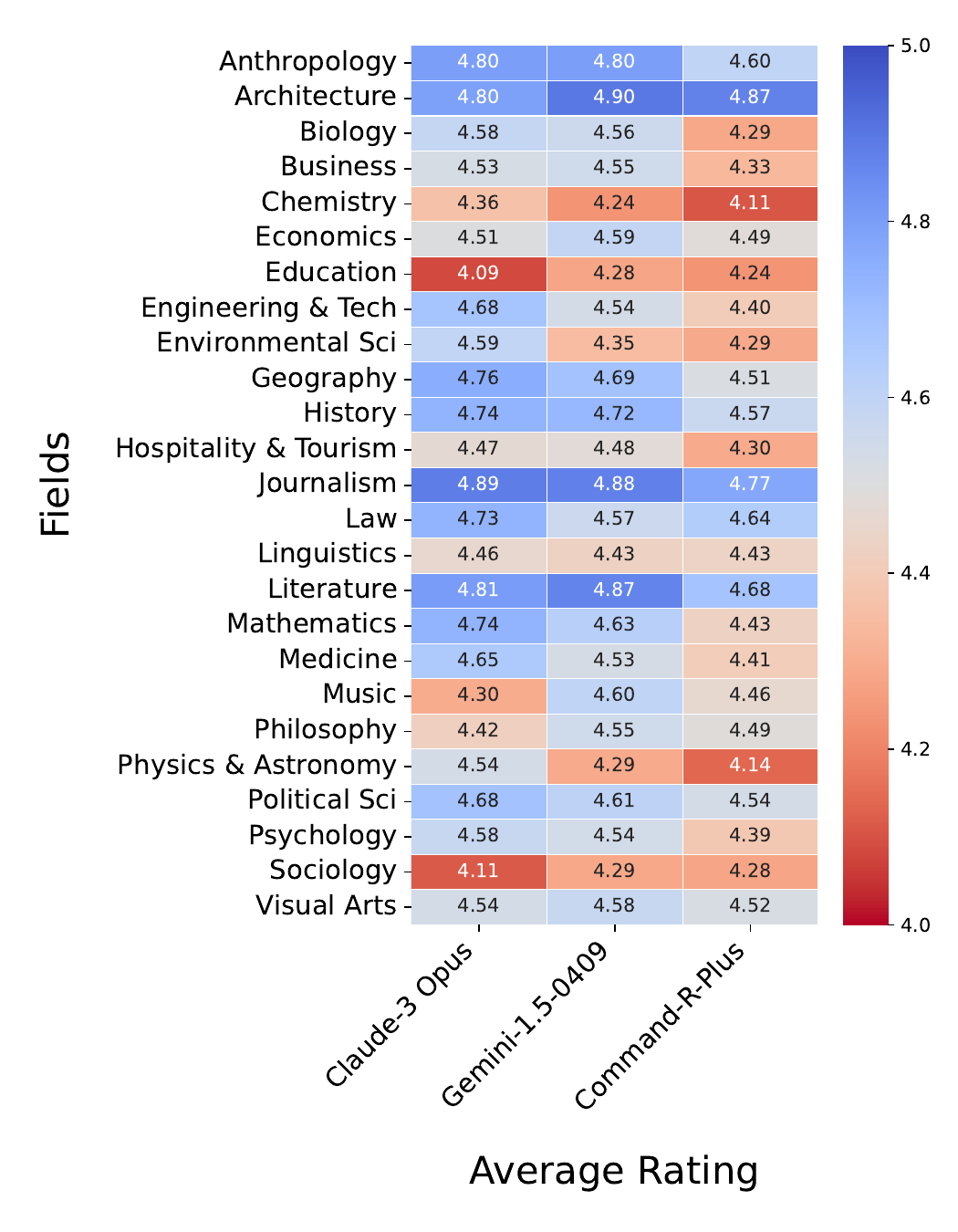}
    \vspace{-25pt}
    \caption{Heatmap of average ratings aggregated by field for Claude-3 Opus, \mbox{Gemini-1.5-0409} and Command-R-Plus.}
    \vspace{-15pt}
    \label{fig:scores_by_field}
\end{figure}

\paragraph{Results by Field.}
 Figure~\ref{fig:scores_by_field} illustrates how model performance
 fluctuates across fields; we show average ratings based on the fine-grained LM evaluation for Claude-3 Opus, \mbox{Gemini-1.5-0409} and Command-R-Plus. Across models, we find
 that a few fields have significantly lower average ratings:
 Education, Sociology and Chemistry. Tasks from a subset of these fields are sometimes 
 subjective (e.g., \textit{Create a lesson plan to teach STEM educators}, \textit{Write a summary of findings about the conclusions of a sociology book}), which can make them hard to evaluate. In other cases, outputs can be lacking technical depth for tasks which require more domain expertise (e.g., \textit{Drafting a experimental protocol for a chemical synthesis procedure.}). On the other hand, tasks in fields such as Literature, History, and Journalism have higher average ratings, some of which focus on factual reporting and narratives that may be easier to reason about.

\begin{figure}[ht!]
    \centering
    \includegraphics[width=.5\textwidth]{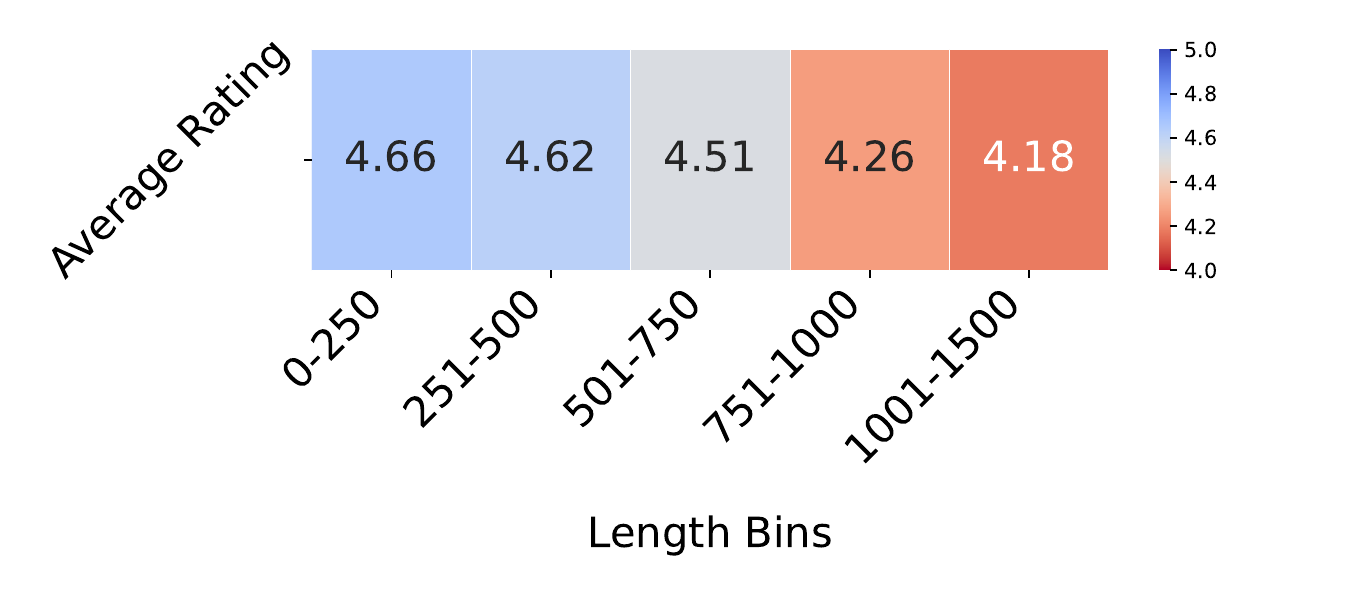}
    \vspace{-25pt}
    \caption{Heatmap of average ratings of Claude-3 Opus aggregated by length (in tokens) of reference output.}
    \vspace{-10pt}
    \label{fig:scores_by_length}
\end{figure}


\paragraph{Results by Length.}

Next, we evaluate whether output length is correlated
with performance. Specifically, we show average ratings based on the fine-grained LM evaluation
for Claude-3 Opus split into bins by length of the reference outputs. Scores stratified by length bins are shown in
Figure~\ref{fig:scores_by_length}. We find that examples which require
longer outputs are significantly harder for models, supported by the fact that examples with longer outputs have lower average scores.




\paragraph{Error Analysis.}

We analyze generated outputs from 3 high-performing models (Gemini-1.5-0409, GPT-4 and Claude-3 Opus) for examples where average ratings of responses are lowest. Broadly, we observe the following patterns:

\begin{itemize}[noitemsep, leftmargin=*, label=\small{$\bullet$}]
    \item \textbf{Lacking depth} (Table~\ref{tab:error_lack_detail}): Writing technical documents requires depth and focus, which was sometimes found to be lacking. For example, concrete statistical results and method details were absent from a task on \textit{writing up a report on an experimental study in clinical psychology}.
    \item \textbf{Verbosity} (Table~\ref{tab:error_verbosity}): A common characteristic of some model outputs was their verbosity. Common patterns included defining jargon  when not necessary, and generating many filler statements that do not introduce new information.
    \item \textbf{Missing information} (Table~\ref{tab:error_missing_info}): There were a few cases where a single output section required multiple pieces of information, but the model entirely missed producing a subset of them.
\end{itemize}

\section{Related Work}

\paragraph{Domain-Specific NLP Benchmarks.} The use of language
technologies in domain-specific scenarios has the potential to help
experts. Prior work has evaluated models through domain-specific
benchmarks for standard tasks like QA \citep{hendryckstest2021,malaviya2024expertqa} and
summarization \cite{hayashi2021wikiasp}. Many benchmarks have been
proposed for specific fields \citep{Rein2023GPQAAG,xia2024fofo},
including law
\citep{shen2022multi,niklaus-etal-2023-lextreme,guha2024legalbench}
and medicine
\citep{tsatsaronis2015overview,pampari-etal-2018-emrqa,jin2019pubmedqa,jin2020disease,fleming2023medalign}.

A notable difference between this line of work and \textsc{Dolomites}
is the task formulation (i.e., QA vs methodical tasks). QA involves
addressing a specific information need in response to a query while
conducting methodical tasks requires following a structured and
consistent procedure, involving multiple steps, to complete a
goal-oriented task.


\paragraph{Naturalistic Evaluation.} Evaluation that is grounded in
realistic use cases, is important for reliable benchmarking \cite{rolnick2024application}. 
Prior efforts on creating NLP benchmarks which are
representative of real user needs use
query logs \citep{nguyen2016ms,kwiatkowski-etal-2019-natural} and chat histories
\citep{wildbench2024}. While these benchmarks are useful
for evaluating responses to generic user queries, they do not
allow us to study their abilities in assisting with
domain-specific tasks in an isolated manner. For instance,
\citet{ouyang2023shifted} find that user requests in chat
histories often involve "planning" and "design", but these are largely
ignored in benchmarks. 
Our work fills this gap by presenting a typology of domain-specific tasks grounded in
realistic scenarios.

\paragraph{Language Models as Writing Assistants.} Recent work has
investigated the potential of language models to act as writing
assistants for domain experts
\citep{calderwood2020novelists,lee2022coauthor,gero2022sparks,li2024value}. While
there is favorable evidence that they can improve productivity of
experts \citep{noy2023experimental}, their usage has broader societal
consequences, including potential impact on the workforce
\citep{eloundou2023gpts}. We analyze a subset of these societal implications relevant
for our methodical tasks in Section~\ref{sec:data}.

\section{Discussion}

As part of \textsc{Dolomites}, we present various data artifacts that can be used to study the abilities of language models in assisting domain experts with writing tasks. We outline these artifacts and their intended use below.

\paragraph{Task Collection.} We present a collection of 519 writing tasks spanning 25 fields that is representative of work undertaken by domain experts on a regular basis. We believe this is the first collection of tasks built with input from domain experts about scenarios in which language models can be useful to them. These tasks can be used to identify applications of LMs in various fields and to cater model development to ecologically valid tasks.

\paragraph{Task Validation Labels.} We conduct an independent assessment of the validity of the tasks and the societal implications of using LMs as writing assistants for these tasks. We believe that these validation labels can be useful to study the practical benefits of using LMs as writing assistants for these tasks and to select tasks that are representative of real-world use. Labels concerning societal implications can enable researchers to take into account important considerations such as anonymity of user data, bias in decision making, accessibility requirements, etc, when considering deployment of language models for assisting experts. 

\paragraph{Task Examples.} Finally, we present examples of tasks that are semi-automatically generated by seeking input from the same experts who provided the tasks. The examples of tasks are meant to represent concrete and plausible instances of the task, so that models can be evaluated on the tasks. To conduct such an evaluation, models are required to generate the output of an example given the task description and the example input. In section~\ref{subsubsec:lm_eval}, we present results on \textsc{Dolomites} using LM-based evaluators as well as other metrics. We find that LM-based evaluators correlate best with human judgements and propose future work to consider the following modes of evaluation on \textsc{Dolomites}:   

\begin{enumerate}[noitemsep, leftmargin=*]
    \item \textbf{Pairwise preference judgements for overall response quality}: Pairwise evaluations from models (especially from Claude-3 Opus) have moderately high agreement with human labels (67\% with ties, 77\% without ties) that are better than other evaluation metrics. These can be used to get an estimate of overall model abilities.
    \item \textbf{Finer-grained LM evaluation for absolute judgements on given axes}: In section \ref{sec:results}, we present fine-grained evaluation on axes that are important for our tasks. We propose future work to conduct finer-grained evaluations on the same axes to gather better insights about the strengths and weaknesses of models.
\end{enumerate}

We believe that as LM-based evaluators improve, we will be able to more accurately evaluate outputs for examples in \textsc{Dolomites}.

\section{Conclusion}

We introduce \textsc{Dolomites}, a benchmark that is
closely tied to realistic use cases of domain experts. The generalization of these use
cases as methodical tasks provides a way to study capabilities of
language models across tasks and domains. We consider a scenario where
AI systems can act as tools for experts to amplify their
problem-solving capabilities \citep{engelbart2023augmenting} and
perform their tasks more efficiently.  We verify that our tasks are
representative across fields and that human oversight is necessary if language 
models propose initial outputs for these tasks.
Evaluation of a broad range of contemporary language models suggests
that there is a large room for models to improve on generating outputs
for our tasks.

Future directions are many and varied. The tasks in \textsc{Dolomites} constitute a mere sample from 25~fields in English
language. We hope to further expand the set of tasks to cover a wider
range of scenarios and languages. We could also consider tasks that
involve modalities other than text in input or output, and multi-turn
settings, where models continually improve their outputs through
feedback and revision. On the modeling front, we will consider sophisticated generation techniques such as the one proposed by \citet{narayan2023conditional}, that first generate a plan of the output and then fill in different sections, potentially with attributions to sources \cite{fierro2024learning}. Our experimental results revealed that
automatic evaluation of generated text is particularly
challenging. Our data contains a single reference output for an
example input and does not model diverse perspectives
of experts and the innate subjectivity of tasks \citep{ganguli2023challenges}. While conventional metrics do not account for this subjectivity, it is
unclear if LM-based evaluators innately capture this
subjectivity. More research is needed to ensure
language model responses are given credit for alternative, but valid
responses.





\section*{Acknowledgments}

We are grateful to the anonymous TACL reviewers and the action editor Minlie Huang for their helpful feedback. We thank the 266 annotators who participated in our data collection studies and the following people for their help with facilitating data collection: Shereen Ashraf, Michelle Chen Huebscher and Michael Sharman. We also thank the following people for helpful discussions and feedback: Sumit Asthana, Tuhin Chakrabarty, Michael Collins, Sunipa Dev, Jacob Eisenstein, Sebastian Gehrmann, Kalpesh Krishna, Tom Kwiatkowski, Matthew Lamm, Kenton Lee, Joshua Maynez, Slav Petrov, Hannah Rashkin, David Reitter, Marco Tulio Ribeiro, Elizabeth Sieber and Kristina Toutanova.

\bibliography{tacl2021}
\bibliographystyle{acl_natbib}

\clearpage

\appendix

\section{Annotation Details}
\label{app:annotation}

\paragraph{Participants.} We recruited 266 participants for our study from Prolific. Participants were required to be fluent in English, and came from 25 different countries, across Africa, Europe, North and South America. In terms of their background, participants were required to have an undergraduate degree and 3 years of work experience in their respective field. These requirements were first enforced through Prolific's audience filters, followed by a screening where participants were asked to self-report their educational qualifications and work experience.  They each provided two tasks, so for each field, we recruited half the number of the participants as the number of tasks reported in Table~\ref{tab:tasks_all}. Lastly, they were required to have at least 50 prior approved submissions and an approval rate of over 99\%. Participants were informed that their provided data will be used to evaluate large language models in realistic scenarios. We obtained prior consent from all annotators before recruiting them for all studies.

\paragraph{Setup.} Annotators were paid \$20 per hour for their work. For task collection, we allocated 40 minutes to write two tasks and for task validation, we allocated 15 minutes per task. For post-editing examples, we allocated 20 minutes per example.

\paragraph{Edit Types.} In a random sample of 100 examples, we found the following types of edits were made to the examples:
\begin{itemize}
\item Fact Addition (88\%): Addition of new statements to the example.
\item Fact Deletion (20\%): Removing statements from the example.
\item Fact Update (65\%): Updating existing statements with further elaboration of details, or adding of new numbers or references.
\item Stylistic Rewrites (76\%): Simplification, paraphrasing, or improving grammar, spelling or tone of text.
\item Reorganization (23\%): Restructuring of sentences, paragraphs or sections in the example, which may be done to fit the task description.
\end{itemize}

\begin{figure*}[t!]
    \centering
    \includegraphics[width=2\columnwidth,keepaspectratio]{images/interface_validation.pdf}
    \vspace{-10pt}
    \caption{Interface shown to annotators for task validation.}
    \label{fig:interface_task_validation}
\end{figure*}

\begin{figure*}[t!]
    \centering
    \includegraphics[width=2\columnwidth,keepaspectratio]{images/interface_editing.pdf}
    \vspace{-10pt}
    \caption{Interface shown to annotators for example post-editing.}
    \label{fig:interface_post_editing}
\end{figure*}

\paragraph{Annotation Interface Screenshots.} We show screenshots of the annotation interfaces presented to annotators for task validation and example \mbox{post-editing} in Figures~\ref{fig:interface_task_validation} and \ref{fig:interface_post_editing} respectively.

\section{Experimental Details}
\label{app:experiments}

\paragraph{Models.} The specific identifiers for the models evaluated in this work are given in Table~\ref{tab:models}. Open-source models were obtained from the HuggingFace model hub, while proprietary models were obtained through the organization's official APIs.

\paragraph{Generation Configurations.} In all generation tasks, we set the temperature for generation to be~0.1. For both example generation and model evaluation, we sampled a maximum of 4,096 tokens (or the maximum sequence length of the model). 

\paragraph{Prompts.} We provide the prompts used for various components of our work. The prompts used for example creation are given in Tables~\ref{tab:query_formulation_prompt}--\ref{tab:example_generation_prompt}. The prompt used to generate the critique shown to annotators is shown in Table~\ref{tab:critique_prompt}. The prompt used to generate outputs from candidate models is shown in Table~\ref{tab:model_eval_prompt}. Finally, the prompt used for generating LM-based judgements for evaluation are shown in Tables~\ref{tab:lm_eval_prompt} and \ref{tab:lm_eval_prompt_fine}.

\begin{table}[t]
\centering
\scalebox{.8}{
   \setlength\extrarowheight{-4pt}
\begin{tabular}{lccc}
\rowcolor{gray!40}
    \toprule
    \textbf{Model Name}        & \textbf{Identifier}                       \\
    \midrule
        \texttt{Claude-3 Opus}              & \texttt{claude-3-opus-20240229}           \\
        \texttt{Command-R-Plus}             & \texttt{command-r-plus}                   \\ 
        \texttt{Gemini-1.5-Pro}\tablefootnote{Accessed from \url{https://aistudio.google.com/app}}             & \texttt{gemini-1.5-pro-latest}            \\
        \texttt{Gemini-1.5-0409}\tablefootnote{Accessed from \url{https://console.cloud.google.com/vertex-ai/generative}}              & \texttt{gemini-1.5-pro-preview-0409}      \\
        \texttt{Gemini-Pro}                 & \texttt{gemini-pro}                       \\
        \texttt{GPT-3.5-Turbo}              & \texttt{gpt-3.5-turbo}                    \\
        \texttt{GPT-4}                      & \texttt{gpt-4-turbo-preview}              \\
        \texttt{Mistral-Large}              & \texttt{mistral-large-latest}             \\
        \texttt{Mixtral-8x22B}              & \texttt{Mixtral-8x22B-Instruct-v0.1}      \\
        \texttt{Mixtral-8x7B}               & \texttt{Mixtral-8x7B-v0.1}                \\
        \texttt{OLMo-7B-Instruct}           & \texttt{OLMo-7B-Instruct}                 \\
    \bottomrule
\end{tabular}
}
\caption{List of models used in our experiments and their identifiers.}
  \vspace{-5pt}
\label{tab:models}
\end{table}

\begin{table}[!t]
\centering
\scalebox{.7}{
  \setlength\extrarowheight{-4pt}
\begin{tabular}{lccc}
\rowcolor{gray!40}
    \toprule
    \textbf{Model} & \multicolumn{1}{c}{\textbf{GPT-4}} & \multicolumn{1}{c}{\textbf{Claude-3 Opus}} & \multicolumn{1}{c}{\textbf{Gemini-1.5-0409}} \\
    \midrule
    \texttt{Claude-3 Opus} & 47.6 & 52.1 & 49.1 \\
    \texttt{Command-R-Plus} & 33.4 & 44.4 & 37.5 \\
    \texttt{Gemini-1.5-Pro} & 43.2 & 48.4 & 53.6 \\
    \texttt{Gemini-1.5-0409} & 44.6 & \textbf{57.6} & \textbf{63.4} \\
    \texttt{Gemini-Pro} & 19.8 & 23.6 & 25.0 \\
    \texttt{GPT-3.5-Turbo} & 14.1 & 13.3 & 14.4 \\
    \texttt{GPT-4} & \textbf{50.0} & 50.0 & 50.0 \\
    \texttt{Mistral-Large} & 29.4 & 31.1 & 28.8 \\
    \texttt{Mixtral-8x22B} & 22.9 & 26.6 & 18.7 \\
    \texttt{Mixtral-8x7B} & 18.2 & 24.0 & 15.9 \\
    \texttt{OLMo-7B-Instruct} & \hspace{.18cm}2.7 &  \hspace{.18cm}3.5 & \hspace{.18cm}3.0 \\
    \bottomrule
\end{tabular}
}
\caption{Model win rate ($\pm3$) against GPT-4 on the \textsc{Dolomites}
  benchmark using three LM-based autoraters (GPT-4, Claude-3 Opus, and
  Gemini-1.5-PP), with a \textit{length penalty}.}
  \vspace{-10pt}
\label{tab:results_lm_lp}
\end{table}


\newpage

\begin{table*}[t!]
    \centering
    \rowcolors{2}{gray!10}{blue!10}
    \begin{tabular}{|p{\dimexpr0.5\linewidth-2\tabcolsep-\arrayrulewidth\relax}|p{\dimexpr0.5\linewidth-2\tabcolsep-\arrayrulewidth\relax}|}
        \toprule
        \multicolumn{2}{|p{2\columnwidth}|}{\cellcolor{gray!50}\textbf{Pattern: Lack of Detail}} \\
        \midrule
        \multicolumn{2}{|p{2\columnwidth}|}{\cellcolor{gray!25}\raggedright \tiny \texttt{\textbf{\underline{TASK DESCRIPTION}} \linebreak \linebreak \textbf{Task Objective}: Designing an observation plan for different celestial bodies and objects using an infrared telescope \linebreak \linebreak \textbf{Task Procedure}: Observing space can be quite a complicated task to achieve, space contains a lot of different celestial bodies and a variety of objects. Some of these objects emits "special" kind of electromagnetic radiation - radiation that we humans cannot see with our eyes. So in this task we're focusing on the invisible glow these objects emit in the infrared part of the spectrum. Our task is to basically decide which celestial bodies (star, planet, galaxies, nebulas and more) we want to study and investigate considering and taking into account their unique infrared features. We also need to plan what kind of telescope is going to be used in order to successfully achieve that mission and literally "see" what we want to see. \linebreak \linebreak \textbf{Additional Notes}: Planning it correctly can save a lot of time and frustrations, taking in account different information like I presented and learning from mistakes can lead to a successful observation. \linebreak \linebreak \textbf{Input Sections}:  \linebreak \linebreak * \textbf{Scientific/main Goal AND target object}: 1 paragraph, 3-4 sentences. To begin with the planning, we mostly need to understand what is our main objective - so we have to outline the scientific goals we aim to gather from the observation such as understanding atmospheric composition of different bodies, stars/planets life cycle and formation process and more.  \linebreak \linebreak * \textbf{Kind of telescope and wavelength range}: 2 paragraphs, 7-8 sentences. The user should provide detailed information regarding what instrument is being used, which includes technical specifications such as focal length of the telescope and eyepieces, apertures, focal ratios, the type of telescope and also what kind of additional items are being used like filters and cameras to detect that special electromagnetic spectrum. Providing what wavelength the observation is going to be in can surely help, infrared radiation varies in different nano-metrica (NIR, MIR, FIR), each wavelength is good for a certain type of observation. \linebreak \linebreak \textbf{Output Sections}: \linebreak \linebreak * \textbf{Observing conditions}: 1 paragraph, 3-4 sentences. We need to know what conditions the observation is taking place and what location, mostly because this plays a real key factor in terms of how the observation would be carried out. Providing information about things like light pollution (Bortle scale), density of air, outside temperature and more.  \linebreak \linebreak * \textbf{Instrument sensitivity}: 1 paragraph, 2-3 sentences. We need to know what are the \hl{specifications of sensitivity of the instruments being used}. Sensors or cameras can detect so many details in comparison to the human eye, as well as filters of eyepieces.}} \\ 
        \midrule
        \multicolumn{2}{|p{2\columnwidth}|}{\cellcolor{gray!25}\raggedright \tiny \texttt{\textbf{\underline{EXAMPLE INPUT}} \linebreak \linebreak * \textbf{Scientific/main Goal AND target object}: Our scientific goal is to study the formation and evolution of young stars within the Orion Nebula. The Orion Nebula is a vast region of star formation, and by observing it in the infrared, we can penetrate the dust and gas that obscure these young stars in visible light. \linebreak \linebreak * \textbf{Kind of telescope and wavelength range}: We will use the Stratospheric Observatory for Infrared Astronomy (SOFIA), a 2.7-meter infrared telescope mounted on a Boeing 747SP aircraft. SOFIA is capable of observing in the mid-infrared and far-infrared wavelengths, from 5.4 to 25.5 microns. We will specifically target the mid-infrared range (8-13 microns), which is optimal for detecting the warm dust and gas associated with young stellar objects. Additionally, we will employ a medium-resolution spectrograph to obtain detailed information about the composition and physical properties of the observed targets.}} \\ \midrule
        \raggedright \tiny \texttt{\textbf{\underline{REFERENCE OUTPUT}}  \linebreak \linebreak * \textbf{Observing conditions}: The observations will be conducted from an altitude of approximately 45,000 feet, above the majority of Earth's water vapor, which absorbs infrared radiation. The flight will take place during a moonless night to minimize light pollution. The observing site will be chosen to have a low Bortle scale rating, indicating minimal light pollution from artificial sources. \linebreak \linebreak * \textbf{Instrument sensitivity}: \hl{SOFIA's mid-infrared camera, FORCAST, has a sensitivity of approximately 0.1 Jy for a point source in a 10-minute integration time. The spectrograph has a spectral resolution of R ~ 1000}, allowing us to resolve individual spectral lines and determine the chemical composition of the target objects.} & \tiny \raggedright \texttt{\textbf{\underline{Gemini-1.5-0409 OUTPUT}} \linebreak \linebreak * \textbf{Observing conditions}: SOFIA operates at altitudes of 38,000-45,000 feet, above 99\% of the Earth's water vapor, which allows for clear observations in the infrared spectrum. The observations will be conducted during nighttime flights to minimize background radiation from the Sun. Additionally, SOFIA's mobility allows it to choose observing locations with minimal light pollution, typically over remote ocean areas or sparsely populated regions, corresponding to Bortle Class 1 or 2. \linebreak \linebreak * \textbf{Instrument sensitivity}: \hlpink{SOFIA's mid-infrared instruments, such as the FORCAST camera and the FLITECAM spectrograph, are highly sensitive to faint infrared emissions}. They can detect temperature variations as small as a few degrees Kelvin and resolve spatial details down to a few arcseconds. This sensitivity enables the detection and characterization of young stellar objects and their surrounding environments within the Orion Nebula.} \tabularnewline
        \bottomrule
    \end{tabular}
    \caption{Sample response showcasing lack of detail in the generated output.}
    \label{tab:error_lack_detail}
\end{table*}

\begin{table*}[t!]
    \centering
    \rowcolors{2}{gray!10}{blue!10}
    \begin{tabular}{|p{\dimexpr0.5\linewidth-2\tabcolsep-\arrayrulewidth\relax}|p{\dimexpr0.5\linewidth-2\tabcolsep-\arrayrulewidth\relax}|}
        \toprule
        \multicolumn{2}{|p{2\columnwidth}|}{\cellcolor{gray!50}\textbf{Pattern: Verbosity}} \\
        \midrule
        \multicolumn{2}{|p{2\columnwidth}|}{\cellcolor{gray!25}\raggedright \tiny \texttt{\textbf{\underline{TASK DESCRIPTION}} \linebreak \linebreak \textbf{Task Objective}: Write a review of a novel. \linebreak \linebreak \textbf{Task Procedure}: You need to write an extensive review of a novel, you need to summarize and give basic facts of the writer and novel, but also give value judgement. \linebreak \linebreak \textbf{Additional Notes}: nan \linebreak \linebreak \textbf{Input Sections}: \linebreak \linebreak * \textbf{Title}: only one or several words, this is the title of the book that you will review. \linebreak \linebreak * \textbf{Writer}: only a name, this is the author's name. \linebreak \linebreak * \textbf{Pages}: only a number, the number of pages in the novel. \linebreak \linebreak * \textbf{Introduction}: 1 paragraph, a general introduction to both the author and their work (the title that will be reviewed in the article) \linebreak \linebreak * \textbf{Summary}: 1 paragraph, a specific summary of the book which gives the reader an insight into the story without spoiling major events or the ending; it needs to stay exciting/enticing. \linebreak \linebreak \textbf{Output Sections}:\linebreak \linebreak * \textbf{Analysis}: 2-3 paragraphs, varying analyses of several aspects of the story, this includes content analysis, but also linguistic and semantical analysis; what is meant by certain passages? What are the cultural/societal connotations? In what way does the writer use language and form? \linebreak \linebreak * \textbf{Verdict}: 1 paragraph, what do we think about the things we just analysed? Is it of high quality or low quality? And in what ways? Why?  \linebreak \linebreak * \textbf{Conclusion}: 1 sentence, summarize everything said above and give a final review in only one sentence.}} \\ \midrule
        \multicolumn{2}{|p{2\columnwidth}|}{\cellcolor{gray!25}\raggedright \tiny \texttt{\textbf{\underline{EXAMPLE INPUT}} \linebreak \linebreak * \textbf{Title}: Dead in Long Beach, California \linebreak \linebreak * \textbf{Writer}: Venita Blackburn \linebreak \linebreak * \textbf{Pages}: 157 \linebreak \linebreak * \textbf{Introduction}: A rising literary star, Venita Blackburn, tackles the complexities of grief and unconventional coping mechanisms in her captivating novel, "Dead in Long Beach, California." This poignant tale explores the depths of human sorrow through the story of a young woman grappling with loss. Blackburn delves into the lengths individuals go to find solace, weaving a narrative that is both heartbreaking and thought-provoking. \linebreak \linebreak * \textbf{Summary}: The novel follows the journey of a young woman who, after the sudden death of her brother, stumbles upon an unexpected and ethically dubious method of coping. Posing as her deceased sibling, she immerses herself in his online world, seeking comfort in the digital memories and connections he left behind. As she delves deeper into her brother's virtual existence, she uncovers hidden secrets and confronts the consequences of her actions.}} \\ \midrule
        \raggedright \tiny \texttt{\textbf{\underline{REFERENCE OUTPUT}}  \linebreak \linebreak * \textbf{Analysis}: Blackburn's novel is a profound exploration of the grieving process. Through the protagonist's unconventional coping mechanism, she examines the ways in which technology can both connect and isolate us in times of loss. The author skillfully intertwines the protagonist's emotional turmoil with the digital landscape, creating a thought-provoking commentary on the nature of identity and the boundaries between the living and the dead. \linebreak \linebreak Furthermore, Blackburn's prose is both lyrical and incisive. For instance, her description of the protagonist staring at her brother's empty online profile, "a hollow echo of a life once lived," perfectly captures the profound sense of emptiness and longing she experiences. The novel's exploration of ethical dilemmas, such as the protagonist's manipulation of online identities, is handled with sensitivity and nuance. It leaves readers to ponder the complexities of right and wrong in the face of overwhelming grief, sparking valuable discussions about the boundaries of acceptable coping mechanisms. \linebreak \linebreak * \textbf{Verdict}: "Dead in Long Beach, California" is a powerful and deeply moving novel. Blackburn's ability to capture the raw emotions of grief and the complexities of our digital lives is commendable. Her exploration of unconventional coping mechanisms and the ethical implications they raise is both thought-provoking and timely. \linebreak \linebreak * \textbf{Conclusion}: Venita Blackburn's "Dead in Long Beach, California" is a haunting and unforgettable tale that lingers long after the last page is turned. It serves as a powerful reminder of the enduring strength of human connection and the profound bond that transcends even death.} & \tiny \raggedright \texttt{\textbf{\underline{GPT-4 OUTPUT}} \linebreak \linebreak * \textbf{Analysis:} Venita Blackburn's "Dead in Long Beach, California" is a masterful exploration of grief and identity in the digital age. Through her concise yet powerful prose, Blackburn challenges the reader to consider the boundaries between the self and the other, the living and the dead. The novel's central theme of coping with loss through digital impersonation raises poignant questions about the authenticity of online relationships and the ethics of assuming another's identity posthumously. Blackburn's use of language is both evocative and precise, capturing the protagonist's turmoil and desperation with striking clarity. The narrative structure, alternating between the protagonist's real-life struggles and her online escapades, effectively mirrors the fragmented nature of grief, enhancing the reader's empathy for the protagonist's plight. Furthermore, Blackburn subtly incorporates societal critiques, particularly concerning the internet's role in shaping our perceptions of death and mourning. The novel's setting in Long Beach, California, with its juxtaposition of urbanity and the vastness of the Pacific, further underscores the protagonist's feelings of isolation and her quest for connection in an increasingly disconnected world. \linebreak \linebreak * \textbf{Verdict}: "Dead in Long Beach, California" is a compelling and thought-provoking novel that showcases Venita Blackburn's talent for delving into complex emotional landscapes with nuance and sensitivity. The quality of Blackburn's writing is evident in her ability to weave a deeply personal story with broader societal implications, making the novel both intimate and universally relevant. The ethical dilemmas and emotional depth explored in the book are handled with a deftness that invites readers to reflect on their own experiences of loss and the ways in which technology intersects with our most human experiences. Blackburn's linguistic prowess and her skillful narrative construction make this novel a high-quality literary work that is both engaging and enlightening. It is a poignant reminder of the power of fiction to explore the nuances of the human condition. \linebreak \linebreak * \textbf{Conclusion}: Venita Blackburn's "Dead in Long Beach, California" is a beautifully crafted novel that offers a profound and nuanced exploration of grief in the digital age, making it a must-read for those who appreciate literature that delves into the complexities of the human heart.} \tabularnewline
        \bottomrule
    \end{tabular}
    \caption{Sample response showcasing verbosity in the generated output (note the \textit{Analysis} and \textit{Verdict} sections).}
    \label{tab:error_verbosity}
\end{table*}

\begin{table*}[t!]
    \centering
    \rowcolors{2}{gray!10}{blue!10}
    \begin{tabular}{|p{\dimexpr0.5\linewidth-2\tabcolsep-\arrayrulewidth\relax}|p{\dimexpr0.5\linewidth-2\tabcolsep-\arrayrulewidth\relax}|}
        \toprule
        \multicolumn{2}{|p{2\columnwidth}|}{\cellcolor{gray!50}\textbf{Pattern: Missing Information}} \\
        \midrule
        \multicolumn{2}{|p{2\columnwidth}|}{\cellcolor{gray!25}\raggedright \tiny \texttt{\textbf{\underline{TASK DESCRIPTION}} \linebreak \linebreak \textbf{Task Objective}: Writing a funding application for a medical research project. \linebreak \linebreak \textbf{Task Procedure}: To write a funding application for a research project, you need to start with a clear and concise idea for the research project, identify matching funding sources, understand the funder's priorities. Then, write a detailed project description and prepare a specific budget. Complete the application form provided by the funder and follow all guidelines. Review the application, submit it, and await feedback. \linebreak \linebreak \textbf{Additional Notes}: It is important to clearly define the research idea, understand the funder's priorities, write a detailed project description, prepare a specific budget, and review before submitting. Common mistakes include overlooking eligibility, using jargon, poor structure of the application, and an overwhelming page layout. \linebreak \linebreak \textbf{Input Sections}: \linebreak \linebreak * \textbf{Project Idea}: 1 paragraph. A clear and concise idea for the research project is essential. \linebreak \linebreak * \textbf{Potential Funding Sources}: 1 paragraph. Estimated cost and identification of potential funding sources that align with the research topic. \linebreak \linebreak \textbf{Output Sections}: \linebreak \linebreak * \textbf{Funding application}: 2 paragraphs. \hl{The funding application will include a summary, an introductory section including background and motivation, detailed project description, and detailed budget.} \linebreak \linebreak * \textbf{Timeline}: In addition, a detailed timeline with key milestones is essential to demonstrate preparedness and careful planning for the project.}} \\ \midrule
        \multicolumn{2}{|p{2\columnwidth}|}{\cellcolor{gray!25}\raggedright \tiny \texttt{\textbf{\underline{EXAMPLE INPUT}} \linebreak \linebreak * \textbf{Project Idea}: We aim to develop a novel risk prediction model for asthma exacerbations in children aged 1-6 years. By integrating clinical data, environmental factors, and genetic markers, we will create a comprehensive tool to identify children at high risk of exacerbations, enabling early intervention and improved disease management. \linebreak \linebreak * \textbf{Potential Funding Sources}: The National Heart, Lung, and Blood Institute (NHLBI) offers funding opportunities for research on respiratory diseases, including asthma. We estimate that the project will require a budget of \$250,000 over five years.}} \\ \midrule
        \raggedright \tiny \texttt{\textbf{\underline{REFERENCE OUTPUT}}  \linebreak \linebreak * \textbf{Funding Application}: \linebreak \linebreak \textbf{Summary}: This application proposes a comprehensive study to develop a risk prediction model for asthma exacerbations in children. The model will integrate clinical, environmental, and genetic data to identify high-risk individuals, facilitating early intervention and improved disease management. \linebreak \linebreak \textbf{Introduction}: Asthma is a common chronic disease that affects millions of children worldwide. Exacerbations can be life-threatening and lead to significant healthcare costs. Current prediction methods are inadequate, and there is an urgent need for more accurate tools. This project will address this gap by developing a novel risk prediction model that combines multiple factors to provide a personalized assessment of exacerbation risk. \linebreak \linebreak \textbf{Project Description}: We will recruit 500 children aged 1-6 years with asthma from the Aspen Medical Group and St. Paul Children's Hospital. Data collection will include clinical assessments, environmental monitoring, and genetic testing. Baseline clinical assessments will include use of short-acting beta agonists, previous asthma-related emergency department attendances and hospitalizations, history of lower respiratory tract infections, environmental tobacco exposure. Environmental monitoring will be performed using indoor air-quality sensors. Genetic testing will include the analysis of specific genes linked to asthma. Outcome events will include asthma-related emergency hospitalization. We will use survival models to estimate the hazard ratio associated with key predictors. The model will be validated using an independent cohort of children. \linebreak \linebreak \textbf{Budget}: The total budget for the project is \$250,000 over five years. This includes costs for participant recruitment, data collection, analysis, and dissemination. The budget includes:\linebreak * \$150,000: Staff salaries \linebreak * \$50,000: Patient recruitment data, collection and analysis \linebreak * \$30,000: Equipment and supplies \linebreak * \$20,000: Travel expenses \linebreak \linebreak * \textbf{Timeline}: \linebreak * Year 1: Participant recruitment and data collection: This include three months for participant recruitment and nine months for data collection. \linebreak * Year 2: Model development and validation: This includes nine months for model development and three months for validation. \linebreak * Year 3: Model refinement, implementation, and real-world testing. \linebreak * Year 4: Dissemination of results and exploration of clinical applications: This include three months for preparing and publishing findings, and nine months for exploring clinical applications. \linebreak * Year 5: Project evaluation and future directions: The entire year will be dedicated to project evaluation and planning for future research.} & \tiny \raggedright \texttt{\textbf{\underline{GPT-4 OUTPUT}} \linebreak \linebreak * \textbf{Funding Application}: Our project proposes the development of an innovative risk prediction model for asthma exacerbations in children aged 1-6 years, a critical period for asthma management and intervention. By leveraging a multidisciplinary approach that combines clinical data, environmental factors, and genetic markers, our model aims to revolutionize how healthcare providers identify children at high risk of asthma exacerbations. This early identification will facilitate timely interventions, potentially reducing the frequency and severity of exacerbations and improving overall disease management and patient outcomes. \linebreak \linebreak The project will unfold in several phases, beginning with the collection and analysis of clinical data from pediatric asthma patients. Concurrently, we will conduct a comprehensive review of environmental factors known to influence asthma exacerbations, such as air quality and allergen exposure, and perform genetic analyses to identify markers associated with increased risk. This integrated data will form the basis of our predictive model, which will be rigorously tested and validated in clinical settings. To support this ambitious project, we are seeking funding of \$250,000 from the National Heart, Lung, and Blood Institute (NHLBI), an organization that prioritizes research on respiratory diseases, including asthma. This budget will cover costs associated with data collection, analysis, personnel, and dissemination of findings, ensuring a thorough and impactful study. \linebreak \linebreak * \textbf{Timeline}: \linebreak \linebreak Year 1: \linebreak - Q1-Q2: Project initiation, including team assembly and training, and development of data collection protocols. \linebreak - Q3-Q4: Begin data collection on clinical data and environmental factors, and initiate genetic marker identification. \linebreak \linebreak Year 2: \linebreak - Q1-Q2: Continue data collection and start preliminary data analysis. \linebreak - Q3-Q4: Development of the initial model based on collected data and initial testing. \linebreak \linebreak Year 3: \linebreak - Q1-Q2: Refinement of the model based on initial testing feedback and integration of additional data as necessary. \linebreak - Q3-Q4: Begin large-scale validation of the model in clinical settings. \linebreak \linebreak Year 4: \linebreak - Q1-Q4: Continued validation and refinement of the model, including adjustments based on ongoing feedback from clinical applications. \linebreak \linebreak Year 5: \linebreak - Q1-Q2: Finalization of the model and preparation of materials for dissemination. \linebreak - Q3-Q4: Dissemination of findings through publications, presentations at conferences, and direct outreach to healthcare providers. \linebreak \linebreak This timeline demonstrates our commitment to careful planning and preparedness, ensuring that each phase of the project builds upon the last towards the successful development and implementation of our novel asthma risk prediction model.} \tabularnewline
        \bottomrule
    \end{tabular}
    \caption{Sample response showcasing missing information in the generated output (note the \textit{Funding application} section).}
    \label{tab:error_missing_info}
\end{table*}

\begin{table*}[t!]
    \centering
    \footnotesize
    \rowcolors{2}{gray!15}{gray!15}
    \begin{tabular}{p{2\columnwidth}}
        \toprule
        \textbf{Query Formulation Prompt} \\
        \midrule
        \raggedright \texttt{Generate 10 search queries for finding specific examples of the given task from the specified field. The search queries should be brief and request documents in more specific contexts than the given task. We would like the documents to contain real examples of the task. List the queries and nothing else. \linebreak \linebreak FIELD: Visual Arts (Graphic design) \linebreak TASK: The objective of this task is to write a catalog entry for an art exhibition. \linebreak QUERIES: 1) Example of catalog entry for art exhibition \linebreak 2) Catalog entry art exhibition Dali \linebreak 3) Notable art catalog entries 2023 \linebreak 4) memorable art catalog entries 2000s \linebreak 5) catalog entry for jackson pollock painting \linebreak 6) frida kahlo painting catalog entry \linebreak 7) picasso guernica catalog entry \linebreak 8) da vinci mona lisa catalog entry \linebreak 9) The Great Wave off Kanagawa catalog entry \linebreak 10) renaissance art exhibition catalog entry \linebreak \linebreak FIELD: {[FIELD}] \linebreak TASK: {[TASK}] \linebreak QUERIES:}
    \end{tabular}
    \caption{Prompt used for generating specific search queries for a task.}
    \label{tab:query_formulation_prompt}
\end{table*}

\begin{table*}[ht!]
    \centering
    \footnotesize
    \rowcolors{2}{gray!15}{gray!15}
    \begin{tabular}{p{2\columnwidth}}
        \toprule
        \textbf{Domain Name Prompt} \\
        \midrule
        \raggedright \texttt{List 20-30 URLs to domain names which will be useful to find real examples for the given task. These websites should be reliable, trustworthy and authoritative sources for an expert in the field. They should be ranked by their likely usefulness. \linebreak \linebreak FIELD: Engineering and Technology (NLP research) \linebreak TASK:Summarizing related work on an NLP subproblem. \linebreak URLs: 1) arxiv.org \linebreak 2) aclweb.org \linebreak 3) ldc.upenn.edu \linebreak 4) nlp.stanford.edu \linebreak 5) aclanthology.org \linebreak 6) towardsdatascience.com \linebreak 7) semanticscholar.org \linebreak 8) openreview.net \linebreak 9) medium.com \linebreak 10) nature.com \linebreak 11) transacl.org \linebreak 12) cambridge.org \linebreak 13) iclr.cc \linebreak 14) aaai.org \linebreak 15) academic.microsoft.com \linebreak 16) nips.cc \linebreak 17) onlinelibrary.wiley.com \linebreak 18) link.springer.com \linebreak 19) naacl.org \linebreak 20) plos.org \linebreak \linebreak FIELD: {[FIELD}] \linebreak TASK: {[TASK}] \linebreak URLs:} \linebreak
    \end{tabular}
    \caption{Prompt used for searching for authoritative domain names for a task.}
    \label{tab:domain_name_prompt}
\end{table*}

\begin{table*}[ht!]
    \centering
    \footnotesize
    \rowcolors{2}{gray!15}{gray!15}
    \begin{tabular}{p{2\columnwidth}}
        \toprule
        \textbf{Example Generation Prompt} \\
        \midrule
        \raggedright \texttt{You are given a description of a task from the field of [FIELD] by an expert. Generate a concrete example of all the Input Sections and Output Sections listed for the given TASK DESCRIPTION. The example should resemble a real example that is written by an expert in the field, and should be highly technical and detailed. \linebreak \linebreak Further instructions:\linebreak - You are also given CONTEXT in the form of Passages from web documents and will need to generate an example based on this CONTEXT. Make sure to generate the example based on the provided CONTEXT. If the CONTEXT is insufficient, you can say "The context is insufficient". \linebreak - Make sure the length of each section matches the required length and the section headers are exactly the same. \linebreak - The example should be highly detailed, and not be generic and vague.\linebreak \linebreak ====CONTEXT====\linebreak \linebreak {[CONTEXT}]\linebreak \linebreak ====TASK DESCRIPTION====\linebreak \linebreak {[TASK DESCRIPTION}]\linebreak \linebreak ====EXAMPLE====} \linebreak
    \end{tabular}
    \caption{Prompt used for generating initial examples for a task.}
    \label{tab:example_generation_prompt}
\end{table*}

\begin{table*}[ht!]
    \centering
    \footnotesize
    \rowcolors{2}{gray!15}{gray!15}
    \begin{tabular}{p{2\columnwidth}}
        \toprule
        \textbf{Critique Generation Prompt} \\
        \midrule
        \raggedright \texttt{You are an expert in the field of [FIELD]. You are given a task description of a writing task from your field and an imperfect example for this task, where an example is a concrete sample of the task. You need to describe what is lacking in the example for the task. You are given a list of properties based on which you should critique the example:\linebreak \linebreak* Inconsistencies: Are there any inconsistencies in the information provided across the input and output?\linebreak * Factual Inaccuracies: Are there any factual inaccuracies in the information presented in the input or how the output is inferred? \linebreak * Structure: Are there any issues with how closely the example follows the instructions specified in the task description? This includes information requested in the task but missing in the example, or mismatch in the length required for a section.\linebreak * Depth: How could the example benefit from more detail? Note that the example should resemble what an expert might write and so it should not be vague with details.\linebreak \linebreak====TASK DESCRIPTION==== \linebreak \linebreak {[TASK DESCRIPTION}] \linebreak \linebreak====EXAMPLE==== \linebreak \linebreak{[EXAMPLE}] \linebreak \linebreak====Critique====} \linebreak
    \end{tabular}
    \caption{Prompt used for generating critiques for model-generated examples.}
    \label{tab:critique_prompt}
\end{table*}

\begin{table*}[ht!]
    \centering
    \footnotesize
    \rowcolors{2}{gray!15}{gray!15}
    \begin{tabular}{p{2\columnwidth}}
        \toprule
        \textbf{Output Generation Prompt} \\
        \midrule
        \raggedright \texttt{You need to perform a writing task from the field of [FIELD]. You are given (1) a task description which contains input and output sections, and (2) an example input for this task, which is a sample of the input sections of the task with concrete details. You need to generate the output sections for the given example input. \linebreak \linebreak - Make sure the length of each output section matches the required length and the section headers are exactly the same. \linebreak - Make sure the output follows the structure of the output sections in the task description, is factually accurate and detailed.\linebreak \linebreak ====TASK DESCRIPTION====\linebreak \linebreak{[TASK DESCRIPTION}]\linebreak \linebreak====EXAMPLE INPUT====\linebreak \linebreak{[EXAMPLE INPUT}]\linebreak \linebreak====EXAMPLE OUTPUT====} \linebreak
    \end{tabular}
    \caption{Prompt used for generating outputs from candidate models for evaluation.}
    \label{tab:model_eval_prompt}
\end{table*}

\begin{table*}[ht!]
    \centering
    \footnotesize
    \rowcolors{2}{gray!15}{gray!15}
    \begin{tabular}{p{2\columnwidth}}
        \toprule
        \textbf{LM-based Overall Preference Evaluation Prompt} \\
        \midrule
        \raggedright \texttt{You are an expert in the field of [FIELD]. You are given a task description of a writing task from your field. For this task description, you are given an input example, which is a concrete sample of the input sections of this task, as well as the reference output, which is the gold standard output for this input. You will be given two candidate outputs for the input example and you need to judge which output is better by comparing it to the reference output. \linebreak \linebreak First, you should say "**output 1**" if output 1 is better, "**output 2**" if output 2 is better and "**same**", if the two outputs are equivalent in quality (note the stars). Then you should explain why you picked this output. \linebreak \linebreak **Important: Keep in mind that longer outputs are not necessarily better quality outputs. Being concise is a good quality for outputs.** \linebreak \linebreak ====TASK DESCRIPTION====\linebreak \linebreak {[TASK DESCRIPTION}]\linebreak \linebreak ====INPUT EXAMPLE====\linebreak \linebreak {[EXAMPLE INPUT}]\linebreak \linebreak ====REFERENCE OUTPUT====\linebreak \linebreak {[REFERENCE OUTPUT}]\linebreak \linebreak ====EXAMPLE OUTPUT 1====\linebreak \linebreak {[EXAMPLE OUTPUT 1}]\linebreak \linebreak ====EXAMPLE OUTPUT 2====\linebreak \linebreak {[EXAMPLE OUTPUT 2}]\linebreak \linebreak ====Decision====} \linebreak
    \end{tabular}
    \caption{Prompt used for generating LM-based judgements.}
    \label{tab:lm_eval_prompt}
\end{table*}

\begin{table*}[ht!]
    \centering
    \footnotesize
    \rowcolors{2}{gray!15}{gray!15}
    \begin{tabular}{p{2\columnwidth}}
        \toprule
        \textbf{LM-based Fine-Grained Evaluation Prompt} \\
        \midrule
        \raggedright \texttt{You are an expert in the field of [FIELD]. You are given a task description of a writing task from your field. For this task description, you are given an input example, which is a concrete sample of the input sections of this task, as well as the reference output, which is the gold standard output for this input. You will be given a candidate output for this input example. You need to evaluate the output by comparing it to the reference output. We will also give you a rubric, which should guide your evaluation. You need to rate the output on the rubric on a scale of 1-5. \linebreak \linebreak Here is the rubric based on which you should evaluate the outputs: \linebreak * Adherence to Task Structure: The output should closely follow the instructions specified in the output sections of the task description. The information requested in the task description should be present in the output, and the sections should be the correct length. \linebreak * Factual Accuracy: There should not be any factual inaccuracies or inconsistencies in the output. \linebreak * Depth: The output text should be technically detailed and thorough, so that it resembles how an expert might conduct the task. \linebreak * Completeness: The output text should be complete and contain all the information requested in the task description. \linebreak * Coherence: The output text should flow logically and be easily understandable. \linebreak \linebreak You should produce the final output as a dictionary in precisely this format: "**output: {{"adherence": \_, "accuracy": \_, "depth": \_, "completeness": \_, "coherence": \_}}**", where you should fill in the spaces with ratings. Make note of the ** required to enclose the output dictionary.\linebreak \linebreak ====TASK DESCRIPTION====\linebreak \linebreak {[TASK DESCRIPTION}]\linebreak \linebreak ====INPUT EXAMPLE====\linebreak \linebreak {[EXAMPLE INPUT}]\linebreak \linebreak ====REFERENCE OUTPUT====\linebreak \linebreak {[REFERENCE OUTPUT}]\linebreak \linebreak ====EXAMPLE OUTPUT====\linebreak \linebreak {[EXAMPLE OUTPUT}]\linebreak \linebreak ====Output====} \linebreak
    \end{tabular}
    \caption{Prompt used for generating LM-based judgements.}
    \label{tab:lm_eval_prompt_fine}
\end{table*}

\end{document}